%% file: main.tex
\pdfoutput=1
\documentclass[10pt,twocolumn]{article}
\usepackage{atlasreport}

\input{arxiv/preamble}

\input{arxiv/macros}
\makeatletter\providecommand\@LN[2]{}\providecommand\@LN@col[1]{}\makeatother

\AtBeginDocument{\hypersetup{pdftitle={ATLAS: Agentic Harness for Whole-Body Intelligence from Human Video},
  pdfauthor={Junli Ren, Kaixiang Yao, Junfeng Long, Tao Huang, Huayi Wang, Yuxiao Chen, Wenhao Ding, Yan Wang, Wenjie Luo, Boris Ivanovic, Ping Luo, Masayoshi Tomizuka, Pieter Abbeel, Marco Pavone, Yuxin Chen, Ran Tian}}}
\runninghead{ATLAS: Agentic Harness for Whole-Body Intelligence from Human Video}

\title{ATLAS: Agentic Harness for Whole-Body Intelligence from Human Video}
\date{November 2026}
\authors{%
  Junli~Ren\affil{*1,3},
  Kaixiang~Yao\affil{*1},
  Junfeng~Long\affil{*1},
  Tao~Huang\affil{3},
  Huayi~Wang\affil{3},
  Yuxiao~Chen\affil{2},
  Wenhao~Ding\affil{2},
  Yan~Wang\affil{2},
  Wenjie~Luo\affil{2},
  Boris~Ivanovic\affil{2\dag},
  Ping~Luo\affil{3\dag},
  Masayoshi~Tomizuka\affil{1\dag},
  Pieter~Abbeel\affil{1\dag},
  Marco~Pavone\affil{2,4\dag},
  Yuxin~Chen\affil{1\ddag},
  Ran~``Thomas''~Tian\affil{1,2\ddag}}
\affiliations{%
  \affitem{1}{UC Berkeley}\quad
  \affitem{2}{Autonomous System \& Physical AI Research Group, NVIDIA}\quad
  \affitem{3}{The University of Hong Kong}\quad
  \affitem{4}{Stanford University}}
\authornotes{\textsuperscript{*}Co-first authors. \textsuperscript{\dag}Equal advising. \textsuperscript{\ddag}Project leads.
  }
\logos{%
  \raisebox{-.5\height}{\includegraphics[height=0.36in]{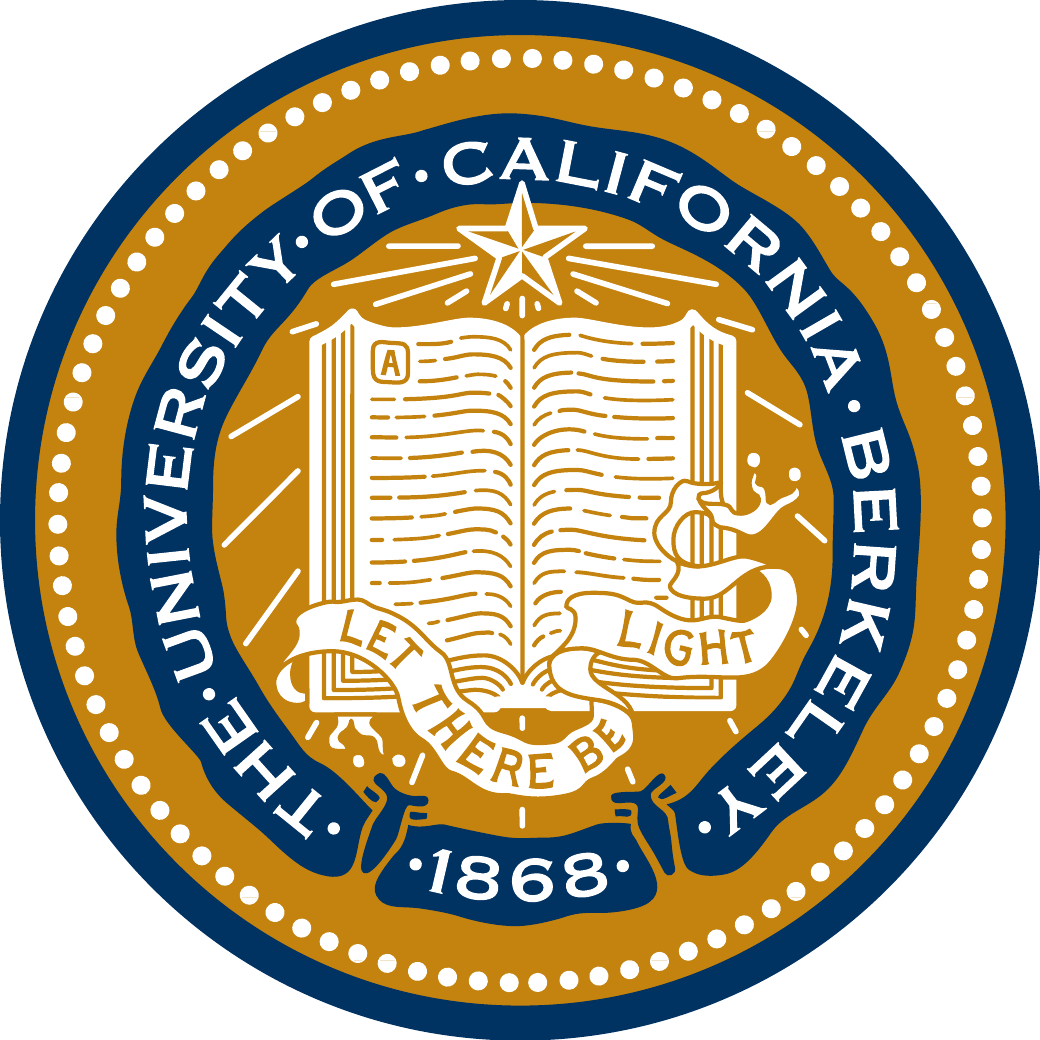}}\hspace{0.22in}%
  \raisebox{-.5\height}{\includegraphics[height=0.19in]{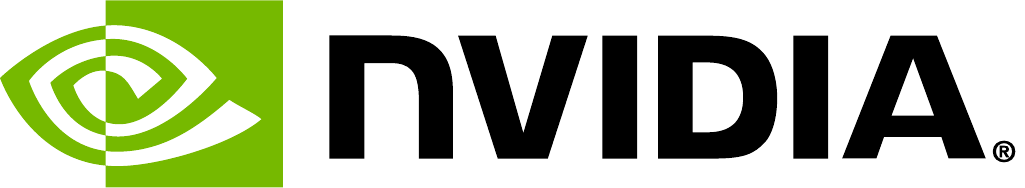}}\hspace{0.22in}%
  \raisebox{-.5\height}{\includegraphics[height=0.38in]{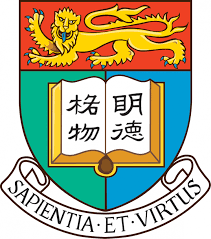}}\hspace{0.22in}%
  \raisebox{-.5\height}{\includegraphics[height=0.3in]{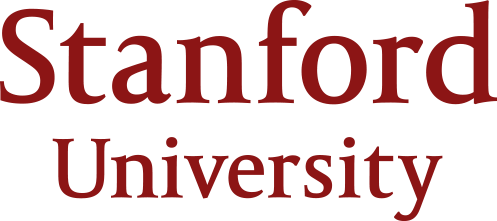}}}
\links{\textcolor{atlasblue}{\bfseries Project page:}~\href{https://atlas-humanoid.github.io}{\nolinkurl{atlas-humanoid.github.io}}}

\renewenvironment{abstract}{\begin{reportabstract}}{\end{reportabstract}}

\begin{document}
\twocolumn[{%
\makereporttitle
\input{arxiv/fig/teaser}
\vspace{6pt}}]
\makereportnotes
\input{arxiv/sec/0_abstract}

\input{arxiv/sec/1_intro}

\input{arxiv/sec/2_related}

\input{arxiv/sec/3_method}
\input{arxiv/sec/4_experiments}
\input{arxiv/sec/5_conclusion}

\input{arxiv/sec/6_ack}

{\linespread{1}\small   %
\bibliographystyle{atlasreport}
\bibliography{main}}

\twocolumn[{\noindent{\sffamily\bfseries\LARGE\color{atlasink}Appendix}\par\vspace{0.7ex}{\color{atlasblue}\hrule height 0.6pt}\vspace{2.2ex}}]
\appendix
\input{arxiv/sec/appendix}

\end{document}

%% file: arxiv/preamble.tex
\usepackage{cuted}
\usepackage{currfile}
\usepackage{caption}

\usepackage{booktabs}
\usepackage{colortbl}
\usepackage{array}
\usepackage{multirow}
\usepackage{makecell}
\usepackage{graphicx}
\usepackage{tikz}
\usetikzlibrary{arrows.meta}
\usepackage{xspace}
\usepackage{pifont}
\usepackage{enumitem}
\usepackage{placeins}
\usepackage{needspace}
\usepackage{xurl}
\usepackage{siunitx}

\newlength{\tabw}
\newcommand{\tabstyle}{\footnotesize\setlength{\tabcolsep}{4pt}\setlength{\tabw}{\linewidth}}

\newcolumntype{P}{S[table-format=3.0]}
\newcolumntype{D}{S[table-format=3.1]}
\newcolumntype{N}{S[table-format=3.0]}
\newcolumntype{R}{S[table-format=3.1]}
\newcolumntype{O}[1]{>{\columncolor{black!8}}#1}
\newcolumntype{B}[1]{>{\columncolor{black!8}[\tabcolsep][34pt]}#1}
\newcolumntype{Z}[1]{>{\columncolor{black!8}[\tabcolsep][0pt]}#1}
\newcolumntype{K}{O{P}}
\newcommand{\up}{\,$\uparrow$}
\newcommand{\down}{\,$\downarrow$}

\makeatletter
\newcommand{\runin}[1]{\expandafter\runin@lc#1}
\def\runin@lc#1{\lowercase{#1}}
\makeatother
\makeatletter
\newcommand{\keyrow}{\noalign{\vbox to\z@{{\color{black!8}\hrule\@height\ht\@arstrutbox\@depth\dp\@arstrutbox\@width\tabw}\vss}}}
\newcommand{\keyrowsep}[1]{\noalign{\vbox to\z@{{\color{black!8}\hrule\@height\dimexpr\ht\@arstrutbox+#1\relax\@depth\dp\@arstrutbox\@width\tabw}\vss}}}
\newcommand{\keyrowstrut}[1]{\rule{0pt}{\dimexpr\ht\@arstrutbox+#1\relax}}
\makeatother

\newlength{\rlstub}

\definecolor{hhOurs}{rgb}{0.21,0.49,0.74}
\definecolor{hhZero}{gray}{0.45}
\definecolor{hhReplay}{gray}{0.75}
\definecolor{hhVLA}{rgb}{0.85,0.37,0.01}
\definecolor{hhLang}{rgb}{0.46,0.44,0.70}
\definecolor{hhRules}{rgb}{0.30,0.55,0.35}
\definecolor{hhVlmA}{rgb}{0.35,0.60,0.65}
\definecolor{hhVlmB}{rgb}{0.75,0.45,0.55}
\definecolor{hhVlmC}{rgb}{0.70,0.60,0.30}
\definecolor{hhVlmD}{rgb}{0.55,0.42,0.32}
\colorlet{hhKey}{black}
\colorlet{hhRef}{black!40}

\newcommand{\TODO}[1]{\textbf{\color{red}[TODO: #1]}}

\usepackage{refcount}
\AtEndDocument{\typeout{MAIN-TABLES p.\getpagerefnumber{tab:main}/\getpagerefnumber{tab:agent}/\getpagerefnumber{tab:rl};
  SECS p.\getpagerefnumber{sec:main}/\getpagerefnumber{sec:agent}/\getpagerefnumber{sec:rl}}}
\newif\ifbudgetdraft
\newcount\budgetowed
\newcount\budgetnow
\newbox\budgetbox
\makeatletter
\newcommand{\budgetfill}[1]{\count@=#1\relax\loop\ifnum\count@>\z@\penalty\z@\vbox to\baselineskip{}\advance\count@\m@ne\repeat}
\makeatother
\newcommand{\TODOlines}[2]{\TODO{#2}%
  \setbox\budgetbox\vbox{\hsize\linewidth\noindent\TODO{#2}}%
  \budgetnow=\numexpr(\ht\budgetbox+\dp\budgetbox)/\baselineskip\relax
  \ifnum\budgetnow<#1\relax
    \global\advance\budgetowed by \numexpr#1-\budgetnow\relax
    \ifbudgetdraft\vadjust{\budgetfill{\numexpr#1-\budgetnow\relax}}\fi
  \fi
  \typeout{BUDGET l.\the\inputlineno: target #1, now \the\budgetnow}}
\newlength{\teaserFinalH}
\newlength{\teaserH}
\ifbudgetdraft\else\fi

\usepackage{etoolbox}
\AtBeginEnvironment{thebibliography}{\raggedright\interlinepenalty=10000\relax}
\usepackage{microtype}

\makeatletter
\renewcommand{\paragraph}[1]{\par\if@nobreak\else\vspace{.4em plus .2em minus .1em}\fi\noindent\textbf{#1.}}
\makeatother
\makeatletter
\def\@textbottom{\vskip\z@\@plus 60pt\relax}
\makeatother

%% file: arxiv/macros.tex
\newcommand{\method}{ATLAS\xspace}
\newcommand{\Sys}[1]{System~#1\xspace}

\newcommand{\timeoutMin}{3}

\newcommand{\nConfigKeep}{3}

\newcommand{\robotDoF}{29}
\newcommand{\controlRateHz}{50}

\newcommand{\rlBoundXHiCm}{12}
\newcommand{\rlBoundYCm}{15}
\newcommand{\rlBoundZCm}{5}
\newcommand{\rlRateMm}{5}
\newcommand{\rlHoldS}{0.5}
\newcommand{\rlRampFrames}{10}
\newcommand{\rlStateDim}{62}
\newcommand{\rlModelInDim}{52}
\newcommand{\rlNeedLo}{5}
\newcommand{\rlNeedHi}{10}
\newcommand{\rlMembers}{5}
\newcommand{\rlHidden}{64}
\newcommand{\rlHistPrev}{9}
\newcommand{\rlLookSlots}{6}
\newcommand{\rlModelSteps}{300}
\newcommand{\rlActorSteps}{200}
\newcommand{\rlLrModel}{$10^{-3}$}
\newcommand{\rlLrActor}{$3{\times}10^{-4}$}
\newcommand{\rlSmoothW}{0.1}
\newcommand{\rlErrUnitCm}{2}
\newcommand{\rlExploreFrac}{0.4}
\newcommand{\rlExploreMm}{2}
\newcommand{\rlWarmup}{5}
\newcommand{\rlArmZeroP}{0.33}
\newcommand{\rlArmMeanP}{0.33}

\newcommand{\simLiftCm}{5}

\newcommand{\mainTextPages}{8}

\newcommand{\slowRowsShown}{12}
\newcommand{\slowFrames}{2}

\newcommand{\judgeModel}{Qwen3-VL-4B\xspace}
\newcommand{\segmenterName}{SAM~3}

\newcommand{\cutTableStepS}{0.04}
\newcommand{\cutFrameStepS}{0.2}
\newcommand{\simHandObjCm}{12}

\newcommand{\slowRecallN}{180}

\newcommand{\slowRecallThrHolmP}{0.049}
\newcommand{\slowIndFalseSuccOff}{42}\newcommand{\slowIndFalseSuccOn}{14}
\newcommand{\slowIndN}{103}\newcommand{\slowIndNotHeldN}{72}\newcommand{\slowIndHeldN}{31}
\newcommand{\slowIndFalseSuccOffK}{30}\newcommand{\slowIndFalseMissOff}{29}\newcommand{\slowIndFalseMissOffK}{9}
\newcommand{\slowIndFalseSuccLogit}{7}\newcommand{\slowIndFalseSuccLogitK}{5}\newcommand{\slowIndFalseMissLogit}{68}\newcommand{\slowIndFalseMissLogitK}{21}
\newcommand{\slowIndFalseSuccRows}{10}\newcommand{\slowIndFalseSuccRowsK}{7}\newcommand{\slowIndFalseMissRows}{77}\newcommand{\slowIndFalseMissRowsK}{24}
\newcommand{\slowIndFalseSuccOnK}{10}\newcommand{\slowIndFalseMissOn}{45}\newcommand{\slowIndFalseMissOnK}{14}
\newcommand{\slowIndDisagreeOff}{39}\newcommand{\slowIndDisagreeOn}{24}\newcommand{\slowIndDisagreeHolmP}{0.025}
\newcommand{\slowIndDisagreeLogit}{26}\newcommand{\slowIndDisagreeRows}{31}
\newcommand{\slowLoopOffK}{0}\newcommand{\slowLoopN}{60}
\newcommand{\slowLoopOnK}{37}
\newcommand{\slowLoopAfterK}{37}\newcommand{\slowLoopAfterN}{39}
\newcommand{\agentSpecN}{28}
\newcommand{\agentHaikuSpecK}{21}
\newcommand{\agentSonnetSpecK}{24}
\newcommand{\agentOpusSpecK}{25}
\newcommand{\agentDevN}{four}
\newcommand{\slowRecallOffD}{18.9}
\newcommand{\slowWrongOffD}{1.1}
\newcommand{\slowRecallThrD}{66.1}
\newcommand{\slowWrongThrD}{3.2}
\newcommand{\slowRecallParaD}{75.0}
\newcommand{\slowWrongParaD}{5.4}
\newcommand{\slowRecallOnD}{78.3}
\newcommand{\slowWrongOnD}{1.6}
\newcommand{\agentOpusRecallD}{78.3}
\newcommand{\agentOpusWrongD}{1.6}
\newcommand{\agentSonnetRecallD}{31.7}
\newcommand{\agentSonnetWrongD}{8.1}
\newcommand{\agentHaikuRecallD}{22.2}
\newcommand{\agentHaikuWrongD}{1.1}
\newcommand{\slowLoopAttemptsPerSeed}{20}
\newcommand{\slowLoopSeedOneOffK}{0}
\newcommand{\slowLoopSeedOneOnK}{12}
\newcommand{\slowLoopSeedOneCycles}{2}
\newcommand{\slowLoopSeedOneSwitch}{9}
\newcommand{\slowLoopSeedOneAfterK}{12}
\newcommand{\slowLoopSeedOneAfterN}{12}
\newcommand{\slowLoopSeedTwoOffK}{0}
\newcommand{\slowLoopSeedTwoOnK}{14}
\newcommand{\slowLoopSeedTwoCycles}{1}
\newcommand{\slowLoopSeedTwoSwitch}{6}
\newcommand{\slowLoopSeedTwoAfterK}{14}
\newcommand{\slowLoopSeedTwoAfterN}{15}
\newcommand{\slowLoopSeedThreeOffK}{0}
\newcommand{\slowLoopSeedThreeOnK}{11}
\newcommand{\slowLoopSeedThreeCycles}{2}
\newcommand{\slowLoopSeedThreeSwitch}{9}
\newcommand{\slowLoopSeedThreeAfterK}{11}
\newcommand{\slowLoopSeedThreeAfterN}{12}
\newcommand{\slowLoopCyclesAll}{5}
\newcommand{\slowRunsWord}{four}

\newcommand{\slowLoopSeedsWord}{three}

\newcommand{\slowLoopCycleSLo}{61}
\newcommand{\slowLoopCycleSHi}{69}
\newcommand{\slowLoopAfterMissesWord}{two}
\newcommand{\slowHaikuKeptRunsWord}{three}
\newcommand{\trashItems}{2}
\newcommand{\trashAttempts}{30}
\newcommand{\trashSeedOneOffK}{0}
\newcommand{\trashSeedOneOnK}{2}
\newcommand{\trashSeedOneFirstAt}{7}
\newcommand{\trashSeedOneFloorAt}{14}
\newcommand{\trashSeedOneCycles}{2}
\newcommand{\trashSeedOneKept}{2}
\newcommand{\trashSeedTwoOffK}{0}
\newcommand{\trashSeedTwoOnK}{2}
\newcommand{\trashSeedTwoFirstAt}{7}
\newcommand{\trashSeedTwoFloorAt}{13}
\newcommand{\trashSeedTwoCycles}{2}
\newcommand{\trashSeedTwoKept}{2}
\newcommand{\trashSeedThreeOffK}{0}
\newcommand{\trashSeedThreeOnK}{2}
\newcommand{\trashSeedThreeFirstAt}{7}
\newcommand{\trashSeedThreeFloorAt}{14}
\newcommand{\trashSeedThreeCycles}{2}
\newcommand{\trashSeedThreeKept}{2}
\newcommand{\trashOffK}{0}
\newcommand{\trashOnK}{6}
\newcommand{\trashN}{6}
\newcommand{\trashFirstAt}{7}
\newcommand{\trashFloorAtLo}{13}
\newcommand{\trashFloorAtHi}{14}
\newcommand{\trashCyclesAll}{6}
\newcommand{\trashCyclesKeptAll}{6}

\newcommand{\realBaZeK}{3}\newcommand{\realBaZeN}{20}\newcommand{\realBaZePctD}{15.0}\newcommand{\realBaZeCpMean}{9.5}\newcommand{\realBaZeCpMed}{8.7}
\newcommand{\realBaCoK}{9}\newcommand{\realBaCoN}{20}\newcommand{\realBaCoPctD}{45.0}\newcommand{\realBaCoCpMean}{4.8}\newcommand{\realBaCoCpMed}{4.2}
\newcommand{\realBaRlsK}{9}\newcommand{\realBaRlsN}{20}\newcommand{\realBaRlsPctD}{45.0}\newcommand{\realBaRlsCpMean}{3.5}\newcommand{\realBaRlsCpMed}{3.2}
\newcommand{\realBaFbK}{10}\newcommand{\realBaFbN}{20}\newcommand{\realBaFbPctD}{50.0}\newcommand{\realBaFbCpMean}{2.5}\newcommand{\realBaFbCpMed}{1.9}
\newcommand{\realCaZeK}{7}\newcommand{\realCaZeN}{20}\newcommand{\realCaZePctD}{35.0}\newcommand{\realCaZeCpMean}{8.4}\newcommand{\realCaZeCpMed}{8.0}
\newcommand{\realCaCoK}{11}\newcommand{\realCaCoN}{20}\newcommand{\realCaCoPctD}{55.0}\newcommand{\realCaCoCpMean}{7.1}\newcommand{\realCaCoCpMed}{6.5}
\newcommand{\realCaFbK}{7}\newcommand{\realCaFbN}{20}\newcommand{\realCaFbPctD}{35.0}\newcommand{\realCaFbCpMean}{5.5}\newcommand{\realCaFbCpMed}{4.7}
\newcommand{\realCaRlsK}{10}\newcommand{\realCaRlsN}{20}\newcommand{\realCaRlsPctD}{50.0}\newcommand{\realCaRlsCpMean}{4.8}\newcommand{\realCaRlsCpMed}{3.1}
\newcommand{\realApZeK}{5}\newcommand{\realApZeN}{20}\newcommand{\realApZePctD}{25.0}\newcommand{\realApZeCpMean}{8.6}\newcommand{\realApZeCpMed}{8.4}
\newcommand{\realApCoK}{9}\newcommand{\realApCoN}{20}\newcommand{\realApCoPctD}{45.0}\newcommand{\realApCoCpMean}{4.5}\newcommand{\realApCoCpMed}{3.6}
\newcommand{\realApFbK}{7}\newcommand{\realApFbN}{20}\newcommand{\realApFbPctD}{35.0}\newcommand{\realApFbCpMean}{3.0}\newcommand{\realApFbCpMed}{2.8}
\newcommand{\realApRlsK}{4}\newcommand{\realApRlsN}{20}\newcommand{\realApRlsPctD}{20.0}\newcommand{\realApRlsCpMean}{3.5}\newcommand{\realApRlsCpMed}{2.9}
\newcommand{\rtyAllZeF}{29}\newcommand{\rtyAllZeW}{76}
\newcommand{\realBaMbK}{11}\newcommand{\realBaMbN}{20}\newcommand{\realBaMbPctD}{55.0}\newcommand{\realBaMbCpMean}{3.2}\newcommand{\realBaMbCpMed}{3.2}
\newcommand{\realCaMbK}{16}\newcommand{\realCaMbN}{20}\newcommand{\realCaMbPctD}{80.0}\newcommand{\realCaMbCpMean}{2.9}\newcommand{\realCaMbCpMed}{2.7}
\newcommand{\realApMbK}{11}\newcommand{\realApMbN}{20}\newcommand{\realApMbPctD}{55.0}\newcommand{\realApMbCpMean}{2.6}\newcommand{\realApMbCpMed}{2.4}
\newcommand{\rtyAllMbN}{36}\newcommand{\rtyAllMbF}{67}
\newcommand{\agtTasksN}{7}
\newcommand{\agtHaExec}{7}\newcommand{\agtHaTime}{82}\newcommand{\agtHaBox}{80}\newcommand{\agtHaDoor}{0}
\newcommand{\agtSoExec}{5}\newcommand{\agtSoTime}{25}\newcommand{\agtSoBox}{0}
\newcommand{\agtOpExec}{7}\newcommand{\agtOpTime}{54}\newcommand{\agtOpBox}{75}\newcommand{\agtOpDoor}{65}
\newcommand{\agtQwExec}{6}\newcommand{\agtQwTime}{35}\newcommand{\agtQwBox}{20}
\newcommand{\agtGpExec}{5}\newcommand{\agtGpTime}{14}\newcommand{\agtGpBox}{60}\newcommand{\agtGpDoor}{100}
\newcommand{\agtAgentsWord}{five}
\newcommand{\tbFirstZe}{47.8}\newcommand{\tbTenZe}{65.6}
\newcommand{\tbFirstMb}{72.2}\newcommand{\tbTenMb}{93.3}

\newcommand{\appFirstBin}{2.6}
\newcommand{\appMbLast}{1.3}\newcommand{\appCoLast}{2.1}\newcommand{\appRlsLast}{2.5}\newcommand{\appZeLast}{4.6}
\newcommand{\appMbLateMed}{1.5}
\newcommand{\appCoLateMed}{2.2}
\newcommand{\appRlsLateMed}{2.7}
\newcommand{\appLateCoP}{5\times10^{-4}}
\newcommand{\appMfLast}{3.5}
\newcommand{\offMbFive}{4.3}\newcommand{\offMbTen}{1.6}\newcommand{\offMbTwenty}{0.9}\newcommand{\offMbForty}{0.4}
\newcommand{\offCoFive}{4.3}\newcommand{\offCoTen}{4.2}\newcommand{\offCoTwenty}{4.1}\newcommand{\offCoForty}{4.1}
\newcommand{\offCmd}{9.7}
\newcommand{\ablFullM}{1.44}
\newcommand{\ablNoShM}{1.59}
\newcommand{\ablOneM}{1.71}
\newcommand{\ablNoLookM}{1.83}
\newcommand{\ablNoGoalM}{2.07}
\newcommand{\ablZeroM}{5.33}
\newcommand{\ablNoHist}{2.40}
\newcommand{\ablNoPhase}{2.46}
\newcommand{\ablOffFull}{0.39}
\newcommand{\ablOffNoSh}{0.96}
\newcommand{\ablStandZero}{3.03}\newcommand{\ablStandOurs}{0.96}\newcommand{\ablStandHi}{1.02}\newcommand{\ablStandN}{twelve}

\newcommand{\realAllN}{60}
\newcommand{\realAllZeK}{15}\newcommand{\realAllZePct}{25}
\newcommand{\realAllMbK}{38}\newcommand{\realAllMbPct}{63}
\newcommand{\hFlZero}{10}\newcommand{\hFlFive}{9}\newcommand{\hFlTen}{9}\newcommand{\hFlTwenty}{6}\newcommand{\hFlThirty}{1}
\newcommand{\hStForty}{8}\newcommand{\hStFortyFive}{5}\newcommand{\hStFifty}{7}\newcommand{\hStSixty}{2}
\newcommand{\hDemoStandGraspCm}{57}\newcommand{\hDemoStandSupportCm}{51}

\newcommand{\slowOnAttempts}{7}
\newcommand{\slowOnTime}{2\,min\,23\,s}
\newcommand{\slowOffAttempts}{12}
\newcommand{\slowOffLifted}{3}

\newcommand{\simXmpPsi}{26}
\newcommand{\simXmpGroot}{27}
\newcommand{\simXmpPi}{13}
\newcommand{\simXmpDz}{30}
\newcommand{\simXmpDp}{8}
\newcommand{\simXmpAct}{25}
\newcommand{\simXmbpPsi}{28}
\newcommand{\simXmbpGroot}{9}
\newcommand{\simXmbpPi}{0}
\newcommand{\simXmbpDz}{1}
\newcommand{\simXmbpDp}{0}
\newcommand{\simXmbpAct}{29}
\newcommand{\simXmpAtlas}{29}
\newcommand{\simXmbpAtlas}{28}
\newcommand{\simBpPsi}{30}
\newcommand{\simBpGroot}{20}
\newcommand{\simBpPi}{28}
\newcommand{\simBpDz}{26}
\newcommand{\simBpDp}{24}
\newcommand{\simBpAct}{28}
\newcommand{\simBpAtlas}{30}
\newcommand{\ctlDoorN}{60}\newcommand{\ctlAppleN}{20}
\newcommand{\ctlAnDoor}{60}\newcommand{\ctlAnDoorErrW}{2.44}\newcommand{\ctlAnDoorErr}{0.18}\newcommand{\ctlAnApple}{18}
\newcommand{\ctlMeDoor}{60}\newcommand{\ctlMeDoorErrW}{2.41}\newcommand{\ctlMeDoorErr}{0.60}\newcommand{\ctlMeApple}{13}
\newcommand{\ctlBfmDoor}{59}\newcommand{\ctlBfmDoorErrW}{2.78}\newcommand{\ctlBfmDoorErr}{0.18}\newcommand{\ctlBfmApple}{13}
\newcommand{\ctlSonDoor}{53}\newcommand{\ctlSonDoorErrW}{6.13}\newcommand{\ctlSonDoorErr}{0.97}\newcommand{\ctlSonApple}{5}
\newcommand{\ctlChDoor}{46}\newcommand{\ctlChDoorErrW}{4.84}\newcommand{\ctlChDoorErr}{2.84}\newcommand{\ctlChApple}{0}
\newcommand{\ctlHeDoor}{45}\newcommand{\ctlHeDoorErrW}{7.54}\newcommand{\ctlHeDoorErr}{1.17}\newcommand{\ctlHeApple}{9}
\newcommand{\ctlAnAppleErrW}{1.89}\newcommand{\ctlAnAppleErr}{0.76}\newcommand{\ctlMeAppleErrW}{3.46}\newcommand{\ctlMeAppleErr}{0.59}
\newcommand{\ctlBfmAppleErrW}{1.13}\newcommand{\ctlBfmAppleErr}{0.88}\newcommand{\ctlSonAppleErrW}{3.10}\newcommand{\ctlSonAppleErr}{1.80}
\newcommand{\ctlChAppleErrW}{8.22}\newcommand{\ctlChAppleErr}{7.47}\newcommand{\ctlHeAppleErrW}{6.04}\newcommand{\ctlHeAppleErr}{1.43}
\newcommand{\ctlTrackersWord}{six}\newcommand{\ctlAppleTrackersWord}{five}
\newcommand{\ctlDoorErrWLo}{2.4}\newcommand{\ctlDoorErrWHi}{7.5}\newcommand{\ctlDoorErrLo}{0.2}\newcommand{\ctlDoorErrHi}{2.8}
\newcommand{\ctlAppleSupportCm}{54.5}\newcommand{\ctlAppleAttempts}{45}\newcommand{\ctlAppleDownCm}{10}
\newcommand{\realBinK}{11}\newcommand{\realBinN}{11}\newcommand{\realBoxK}{17}\newcommand{\realBoxN}{20}
\newcommand{\realPickSucc}{\realAllMbK/\realAllN}\newcommand{\realBinSucc}{\realBinK/\realBinN}\newcommand{\realBoxSucc}{\realBoxK/\realBoxN}
\newcommand{\wrAttemptsWord}{ten}
\newcommand{\wrTakeWord}{ten}
\newcommand{\wrProbeWord}{two}
\newcommand{\wrPartialWord}{two}
\newcommand{\wrWholeCountWord}{six}
\newcommand{\wrCovLateFloor}{90}
\newcommand{\wrLettersWord}{five}
\newcommand{\wrLetterRule}{half}
\newcommand{\wrSnapFirst}{2}
\newcommand{\wrSnapLast}{5}
\newcommand{\wrWordWidthCm}{35}
\newcommand{\wrStrokeM}{1.1}
\newcommand{\wrLiveWord}{eight}
\newcommand{\wrBoardMoveMm}{28}
\newcommand{\wrForceN}{2}
\newcommand{\wrGainMm}{0.1}
\newcommand{\wrSeekBelowN}{0.05}
\newcommand{\wrSeekMm}{1}
\newcommand{\wrBackoffAboveN}{4}
\newcommand{\wrRampSteps}{10}
\newcommand{\wrPressMm}{5}
\newcommand{\wrFadeSteps}{2}
\newcommand{\wrFadeLeadSteps}{12}
\newcommand{\wrBoundLoCm}{6}
\newcommand{\wrBoundHiCm}{1}
\newcommand{\wrDepthRate}{0.8}
\newcommand{\wrDepthMaxCm}{3}
\newcommand{\wrSeekBelowJointsN}{0.95}
\newcommand{\wrJointsRmsN}{0.5}
\newcommand{\wrBoardLoop}{83}
\newcommand{\wrErrLoop}{2.5}
\newcommand{\wrBoardJoints}{84}
\newcommand{\wrErrJoints}{2.5}
\newcommand{\wrBoardActorOne}{37}
\newcommand{\wrErrActorOne}{5.2}
\newcommand{\wrBoardActorTwo}{34}
\newcommand{\wrErrActorTwo}{3.4}
\newcommand{\wrErrJointsZero}{15.8}
\newcommand{\wrErrLoopZero}{15.9}

%% file: arxiv/fig/teaser.tex
\begin{center}
\includegraphics[width=\linewidth]{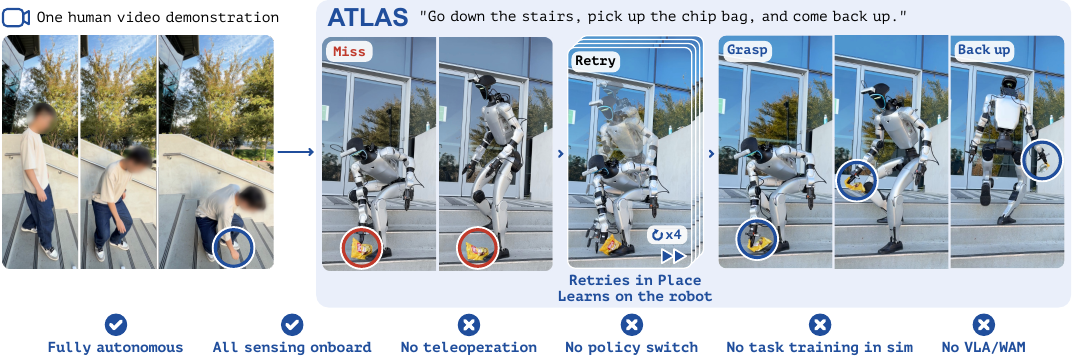}\par
\captionof{figure}{\textbf{\method in the real world.}
From one human video demonstration, with no robot demonstrations, the robot goes down an outdoor staircase and comes back up with the chip bag its prompt names. Its first grasp misses, so it retries in place and grasps the bag, learning on the robot as it goes. No pretrained vision-language or vision-language-action model is fine-tuned.}
\label{fig:teaser}
\end{center}

%% file: arxiv/sec/0_abstract.tex
\begin{abstract}
Scaling humanoid whole-body intelligence is often pursued through large-scale training of vision-language-action (VLA) or world action models (WAMs). 
Yet recent advances in pretrained vision-language agents provide reusable knowledge about objects, task goals, and physical interactions. This motivates acquiring new whole-body skills by adapting human motion references to new tasks via pretrained task reasoning and learning execution corrections online, \textbf{without} large-scale VLA or WAM training.
We present \method (Agentic Test-time Learning and Adaptation System), an agentic harness that edits human motion for humanoid loco-manipulation from one human video demonstration and one prompt. The harness splits the motion into navigation and interaction phases as specified by a pretrained agent, plans the navigation, and aligns each interaction phase with the target object by its affordance and perceived geometry; a frozen whole-body tracker then follows the edited reference. During execution, online model-based RL mitigates the tracking gap between the edited reference and the robot's motion, while a slow system lets the agent refine the harness's grounding of each scene from robot observations. Success checks detect failures and trigger retries in place, generating new execution data and observations for further control and grounding adaptation.
To the best of our knowledge, \method is the first agentic whole-body intelligence system to achieve this without task-specific robot demonstrations, VLA or WAM training, or task-specific simulation training.
We demonstrate \method on a real humanoid performing coordinated whole-body loco-manipulation, including object manipulation on both ground and stairs, force-controlled contacts like writing, using only onboard sensing.
In both simulation and real-world experiments, \method acquires skills from one human video demonstration and one prompt per task and transfers them across objects and scenes.
On real-robot floor picks of three objects, online correction raises success from \realAllZePct\% to \realAllMbPct\%.
On three whole-body pick tasks of the SIMPLE benchmark, a single human video serves all three tasks, and \method succeeds in \simXmpAtlas, \simXmbpAtlas{} and \simBpAtlas{} of 30 episodes with onboard sensing, competitive with VLA, WAM and imitation policies fine-tuned on 100 robot demonstrations per task.

\end{abstract}

%% file: arxiv/sec/1_intro.tex
\section{Introduction}
\label{sec:intro}

Scaling humanoid whole-body intelligence with vision-language-action (VLA) or world action models (WAMs) typically requires substantial robot or human data and compute to map visual observations and task instructions to robot actions~\cite{gr00tn1,ye2026dreamzero}.
Yet pretrained vision-language agents increasingly provide reusable knowledge about objects, task semantics, and physical interactions~\cite{rekep2024,guo2026robodawn}.
They can turn this knowledge into robot programs~\cite{cap2022}, geometric constraints~\cite{rekep2024}, or commands through closed-loop action interfaces~\cite{guo2026robodawn}.
Recent agentic tabletop manipulation systems need no task-specific robot demonstrations and have outperformed several SOTA VLA/WAM baselines on manipulation benchmarks~\cite{chu2026openrua,yuan2026democratizing}.
For humanoids, this motivates enabling pretrained vision-language agents to perform new tasks requiring coordinated whole-body motion, without further large-scale VLA/WAM training or task-specific simulation training.

Human video can provide a whole-body reference for an agent's plan through motion recovery and retargeting~\cite{gvhmr2024,gmr2025}.
The reference must still be grounded in the instructed objects and scene and adapted to the robot's embodiment.
Our key idea is to let a pretrained agent specify reference edits while the robot learns from its attempts to refine grounding and correct the frozen tracker's end-effector commands.
For example, to reuse a banana-picking motion to collect a paper ball, the robot must locate the ball and align the reach with its geometry. Even with the target correctly placed, tracking error can leave the hand short of contact.

We present \method (Agentic Test-time Learning and Adaptation System), an agentic harness for humanoid loco-manipulation from one human video demonstration and one prompt.
\method connects a vision-language agent (\Sys{2}) to a frozen whole-body tracker (\Sys{0}) only through edited motion references, replacing an intermediate policy pretrained on robot data.
Once per task, the agent segments the demonstration into steps and specifies target objects, contact affordances, and success checks.
On each attempt, the harness grounds these affordances in perceived geometry, navigates into position if needed, and aligns the reference's reach with the contact poses (\cref{sec:method,fig:pipeline}).
A progress indicator then applies the success checks, and a miss is retried in place.

\begin{figure}[t]
  \centering
  \includegraphics[width=\linewidth]{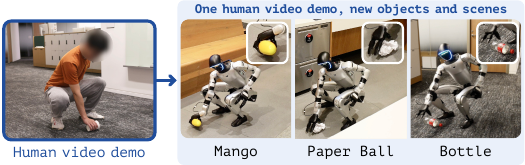}
  \caption{\textbf{One human video demonstration transfers to new objects and scenes.}
  A person picks a cup off the floor.
  On real hardware, the same motion drives the robot to grasp a mango, a paper ball and a bottle in three scenes.
  The agent writes each task's specification, and no pretrained vision-language or vision-language-action model is fine-tuned.}
  \label{fig:one_motion}
\end{figure}

The harness uses execution feedback to adapt both grounding and control.
During execution, its fast system uses online model-based RL to learn the tracker's command-response dynamics from scratch in each session and corrects the end-effector command at every control step during reaching (\cref{sec:fast}).
Outside the control loop, its slow system uses the same agent to revise the perception phrases and the progress indicator from robot observations, retaining only revisions that improve performance when replayed on saved observations (\cref{sec:slow}).
Retries supply further experience for both systems.

Using only onboard sensing, a real humanoid running \method is able to manipulate objects on both ground and stairs (\cref{fig:teaser,fig:tasks}), perform contact-force-controlled tasks like writing letters on a whiteboard (\cref{fig:writing}), and reuse a demonstrated pickup motion across objects and scenes (\cref{fig:one_motion}).
At a fixed apple placement in simulation, our model-based learner achieves about 63\% lower median tracking error during the pre-grasp hold than a model-free actor. Both correct the end-effector command at every control step.
Across real-robot configurations with differing calibrations, our learner achieves \realAllMbPct\% floor-pick success versus \realAllZePct\% without correction (\cref{sec:ablation}).
In held-out replay of robot observations, agent review improves object recall and reduces false success judgments (\cref{sec:exp_slow}).
On the public SIMPLE benchmark, \method reuses one human video across three loco-manipulation tasks without task-specific robot demonstrations, achieving success rates competitive with VLA, world-action-model, and imitation policies fine-tuned on 100 robot demonstrations per task (\cref{tab:benchmarks}).

%% file: arxiv/sec/2_related.tex
\newpage
\section{Related Work}
\label{sec:related}
\paragraph{VLA and WAM for humanoid whole-body intelligence}
Hierarchical stacks such as Helix 02~\cite{figure2026helix02} and Dyna-2.1~\cite{dyna2026dyna21} combine vision-language reasoning (\Sys{2}), learned action generation (\Sys{1}), and whole-body control (\Sys{0}).
Training VLAs~\cite{pi05,gr00tn1} and WAMs~\cite{ye2026dreamzero} to map visual observations and instructions to actions requires substantial data and compute.%
Pretrained agents reuse task knowledge for reasoning and grounding~\cite{cap2022,rekep2024,guo2026robodawn}; manipulation results without task-specific robot demonstrations~\cite{chu2026openrua,yuan2026democratizing} suggest a route to new behaviors without additional task-specific VLA/WAM training.

\paragraph{Whole-body controllers}
Pretrained whole-body controllers provide reusable motor capabilities from motion references or goal commands~\cite{luo2025sonic,bfmzero2025}.
However, embodiment differences and mismatches between training and deployment references can limit tracking accuracy~\cite{ling2026gentrack}.
For loco-manipulation, residual end-effector errors can prevent the robot from reaching the contact targets required for successful interaction.
Keeping ScaleBFM~\cite{zeng2026scalebfm} frozen, \method uses online model-based RL to learn its command-response dynamics from robot attempts and train an actor that corrects end-effector commands during reaching.

\paragraph{Agents and harnesses for humanoids}
Vision-language agents are increasingly used to bridge task reasoning and humanoid control, offering an alternative to VLA-based intermediate policies.
HomeBody~\cite{huh2026homebody} and KINO~\cite{chen2026kino} use agents to select and sequence reusable skills or intermediate whole-body pose goals, leaving motion generation and execution to downstream planning and control modules.
KPI~\cite{wang2026kpi} uses an agent's parameterized wrist references and contact objectives to adapt arm contact behavior online.
HuGo~\cite{choi2026hugo} generates closed-loop policy code and refines it through task-specific simulation rollouts, with further refinement from real-robot feedback.
A remaining challenge is to enable new whole-body task behaviors without handcrafting task-specific skills or requiring task-specific simulation optimization.
\method uses pretrained agent reasoning to adapt human video references for a frozen whole-body tracker, refining grounding and execution online without task-specific robot demonstrations or additional VLA/WAM training.

\paragraph{From human video to humanoid motion}
Human videos provide motion and interaction structure for acquiring new humanoid behaviors.
VideoMimic reconstructs human motion and scene geometry from video, then learns contextual control policies in simulation~\cite{videomimic2025}.
Other methods adapt video-derived references through object-aware retargeting~\cite{li2024okami}, closed-loop posture and contact adjustment~\cite{he2026praxis}, or contact-aware trajectory optimization~\cite{genhoi2026}.
RoboReact further uses a VLM to refine video-derived skill keyframes and interaction parameters from physical calibration rollouts~\cite{he2026roboreact}.
\method builds on reference-based execution by coupling agent-driven grounding refinement for the instructed objects and scene with online model-based RL that corrects end-effector commands within each reach.

\paragraph{On-device learning}
Residual learning augments existing robot controllers and policies with learned corrections~\cite{johannink2019residual,resfit2025,resmimic2025,fang2026find}.
For frozen whole-body controllers, adaptation can operate on motion commands through simulation-based optimization~\cite{xie2026ttco} or offline-learned kinematic and odometry models~\cite{dong2026hero}.
OSRAM fine-tunes a simulation-meta-trained command-response model on hardware and uses MPPI to optimize command sequences online~\cite{huang2026osram}.
At this command interface, \method learns a command-response model and a correction actor from scratch in each deployment session.
Unlike OSRAM, its model needs no meta-training in simulation, and a trained actor replaces online sampling-based optimization; unlike iterative learning control~\cite{bristow2006ilc}, which updates a correction between repetitions, it also corrects the command within each reach.
It fits the model to robot attempts and trains the actor through model rollouts~\cite{deisenroth2011pilco,wu2022daydreamer,liao2026supermimic}, yielding end-effector command corrections at every control step during reaching.

\begin{figure*}[t]
  \centering
  \includegraphics[width=\textwidth]{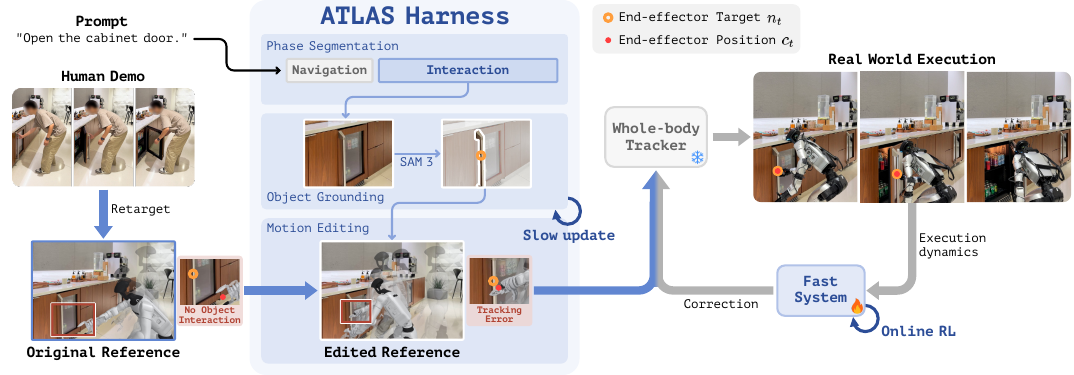}
  \caption{\textbf{\method links a vision-language agent (\Sys{2}) and a frozen whole-body tracker (\Sys{0}) through edited motion references.}
  Left: human video supplies whole-body motion; the agent specifies interaction steps, objects, contacts, and success checks.
  Right: onboard perception grounds contact targets, and the harness aligns the reference for execution.
  The fast system learns end-effector command corrections; the slow system revises what perception is asked to find and the success checks through replay validation.
  Here, a human cabinet-opening video and the prompt ``Open the cabinet door.'' drive the robot to open a cabinet.}
  \label{fig:pipeline}
\end{figure*}

%% file: arxiv/sec/3_method.tex
\section{\method: Agentic Test-time Learning and Adaptation System}
\label{sec:method}

\subsection{Task setup and deployment loop}
\label{sec:overview}
\label{sec:retry}
A related human video already contains the whole-body coordination a task needs; what changes across tasks is mainly the interaction with the object.
\method therefore reuses the demonstrated motion and adapts it to the instructed interaction, even when the objects differ from those in the video (\cref{fig:pipeline}).
On each attempt, the harness locates the instructed object, walks the robot to it if needed, and shifts the demonstrated end-effector target onto it; the frozen tracker executes the edited motion, and a progress indicator checks the end state and retries failures in place.
Two kinds of error remain, and two learners reduce them from the robot's own attempts: the fast system corrects the tracker's end-effector error during execution (\cref{sec:fast}), and the slow system lets the agent revise what perception looks for and how success is judged (\cref{sec:slow}).
An \emph{attempt} ends in success or failure, an \emph{episode} groups the attempts at one object placement, and a \emph{session} keeps experience across episodes.

\subsection{From human video to grounded motion}
\label{sec:motion_reference}
\label{sec:spec}
\paragraph{Motion reference}
GVHMR~\cite{gvhmr2024} recovers the person's motion and GMR~\cite{gmr2025} retargets it to the robot, giving a reusable reference of body poses and end-effector trajectories.
The agent splits it into interaction steps, such as a pick and a place, and marks each step's interaction start, contact, settle, and end frames.

\paragraph{Task specification}
The agent specifies each step only in text and a few parameters, which the slow system can later revise.
From the instruction, measurements of the reference such as the contact height and action axes, and a description of the robot's end effector, it writes an object phrase (and a destination phrase if needed), contact and closure parameters, and a visible end state that the progress indicator checks for success.
For example, when a banana pickup is reused to pick up a paper ball with a fingered hand, it names the paper ball, chooses an \texttt{enclose} closure, and defines success as the ball held above the floor.

\paragraph{Grounding and navigation}
At run time, \segmenterName~\cite{sam3_2025} segments the phrase's \emph{grounding target}, the object or the part to interact with, such as a handle, and depth gives its center, extent, and axes, providing real-time, closed-loop feedback on the object position.
From this geometry and the specified interaction, the harness computes the wrist orientation and the \emph{target contact point} $p^\star$, for example the center of a handle.
Navigation ensures that the interaction starts from the same pose relative to the object as in the retargeted reference: the \emph{interaction station} is derived backward from the reference's interaction, as its pelvis pose at the interaction start relative to its contact point, and a whole-body walking planner~\cite{wang2026motionbricks} generates walking references that the frozen tracker follows to it.

\subsection{Editing the reference and executing it}
\label{sec:edit}
The harness changes only the interacting end effector's target and, for higher objects, the body height, so the rest of the motion keeps the demonstrated whole-body coordination that the frozen tracker can follow.

\paragraph{Placing the reference}
At the \emph{handoff}, the robot stops walking and starts the demonstrated motion from where it stands: the reference is turned and shifted on the floor so that its starting pelvis coincides with the robot's.
To compare targets, measurements, and corrections while the body moves, all positions are then expressed in the \emph{attempt frame}, fixed to the floor at handoff and pointing forward, left, and up; the robot knows where its pelvis is in this frame from leg kinematics, without external localization.
The \emph{contact point} is the part of the end effector meant to touch the object, at a fixed offset from the wrist.

\paragraph{Editing the end-effector target}
Placed this way, the reference still moves the end effector toward where the person's object was, so the harness shifts the end-effector target onto $p^\star$, gradually so that the reach stays smooth.
With $\tau$ the annotated contact frame and $c^{\mathrm{ref}}_t$ the placed reference's contact point at control step $t$, the target is
\begin{equation}
  n_t \;=\; c^{\mathrm{ref}}_t \;+\; \alpha_t\bigl(p^\star - c^{\mathrm{ref}}_\tau\bigr),
  \label{eq:contact_alignment}
\end{equation}
where $\alpha_t$ rises from 0 at \emph{reach onset}, the start of the wrist's final approach to the object, to 1 at $\tau$ in proportion to the wrist path traveled, so that $n_\tau=p^\star$.
New head-camera estimates update $p^\star$ during the crouch, closing the loop on the object position.

\paragraph{Adapting height}
For higher objects, body and arm references rise by the height difference over the reach, with the feet on the floor and the pelvis capped at standing height; for lower objects, only \cref{eq:contact_alignment} lowers the end effector.

\paragraph{Whole-body tracking}
\label{sec:wbc}
One tracker, ScaleBFM~\cite{zeng2026scalebfm}, follows both the walking and the interaction references at \controlRateHz\,Hz without switching policies, and stays frozen during deployment; for stair traversal, we fine-tune it beforehand on general stair-climbing data, not on any task.
For load-bearing grasps, the reference pauses at contact with the end effector open (the \emph{hold}) before closing; pushes, pulls, and presses close at contact.

\subsection{Fast system: command correction via online model-based RL}
\label{sec:fast}
\begin{figure}[t]
  \centering
  \includegraphics[width=\linewidth]{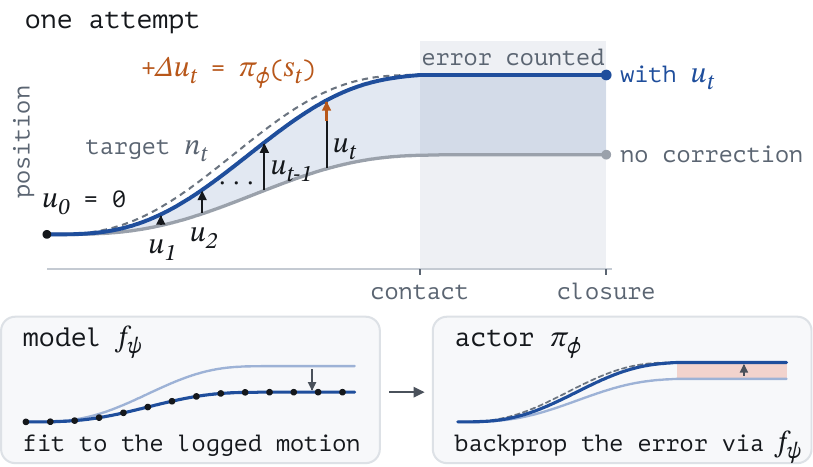}
  \caption{\textbf{The fast system accumulates bounded command corrections to reduce contact error during the hold.}
  In this schematic, the uncorrected tracker undershoots $n_t$; the shaded band shows the shift $u_t$ produces.
  Logged motion trains the dynamics model $f_\psi$, and backpropagating hold error through its rollouts trains the actor $\pi_\phi$.}
  \label{fig:fast}
\end{figure}

Even with correct grounding, the frozen tracker leaves systematic end-effector errors that vary across attempts: its response depends on where the command lies relative to the body and on the posture along the reference.
The fast system learns how the robot responds to end-effector commands with online model-based RL and corrects the command from reach onset through the hold (\cref{fig:fast}).

\paragraph{What the actor controls}
Let $c_t$ be the contact point measured by forward kinematics.
To bring $c_t$ to $n_t$ during the hold, the actor adds an offset $u_t$ on attempt-frame axes to the interacting wrist's position command (the \emph{harness command}); $n_t$ and the tracker's weights stay unchanged.
The actor outputs increments $\Delta u_t=\pi_\phi(s_t)$ of at most \rlRateMm\,mm per control step that accumulate,
\begingroup
\setlength{\abovedisplayskip}{4pt plus 1pt minus 1pt}
\setlength{\belowdisplayskip}{4pt plus 1pt minus 1pt}
\setlength{\abovedisplayshortskip}{0pt}
\setlength{\belowdisplayshortskip}{4pt plus 1pt minus 1pt}
\begin{equation}
  u_t=\operatorname{clip}(u_{t-1}+\Delta u_t),
  \label{eq:actor}
\end{equation}
\endgroup
starting from $u=0$ at reach onset; clipping keeps $u_t$ within fixed bounds.
Because the state $s_t$ (defined below) includes $u_{t-1}$, the actor can still reach any correction within these bounds.

\paragraph{Learning how the robot responds}
After each attempt, a dynamics model learns how the contact point responds to the commands, working relative to the shoulder.
Writing $c_t=b_t+h_t$, where $b_t$ is the interacting arm's recorded shoulder position, it predicts
\begingroup
\setlength{\abovedisplayskip}{4pt plus 1pt minus 1pt}
\setlength{\belowdisplayskip}{4pt plus 1pt minus 1pt}
\setlength{\abovedisplayshortskip}{0pt}
\setlength{\belowdisplayshortskip}{4pt plus 1pt minus 1pt}
\begin{equation}
  \hat h_{t+1}=\hat h_t+f_\psi(z_t),\qquad \hat c_{t+1}=b_{t+1}+\hat h_{t+1},
  \label{eq:model}
\end{equation}
\endgroup
where $z_t$ holds the reference phase (0 at reach onset, 1 at contact), $h_t$ and its last change, and recent and look-ahead commanded contact points relative to $b_t$.
Each command is the harness command plus the applied correction.
We fit an ensemble of such models with randomized priors~\cite{osband2018rpf} and a multi-step prediction loss on all attempts of the session except falls, including failed grasps; rollouts start from a measurement and feed back the predicted $\hat h$.

\paragraph{Training the actor}
The actor solves a finite-horizon RL problem with state $s_t$, action $\Delta u_t$, and transitions given by the robot under the frozen tracker.
With the model fixed, it learns by backpropagating the predicted hold error through the model's reach-and-hold rollouts~\cite{deisenroth2011pilco}.
The actor sees what the model sees plus the target: its state $s_t$ adds to $z_t$ the errors $n_{t+k}-c_t$ for $k=0,\rlNeedLo,\rlNeedHi$, the applied correction, and the  correction not yet applied because of command delay.
Each rollout starts from a recorded attempt's measured reach-onset state and replays its shoulder trajectory and harness command, substituting the actor's corrections under the execution limits; this treats body motion as fixed context, an approximation for bounded corrections around a recorded movement.
For load-bearing grasps, the reward is the contact error during the hold until closure,
\begingroup
\setlength{\abovedisplayskip}{4pt plus 1pt minus 1pt}
\setlength{\belowdisplayskip}{4pt plus 1pt minus 1pt}
\setlength{\abovedisplayshortskip}{0pt}
\setlength{\belowdisplayshortskip}{4pt plus 1pt minus 1pt}
\begin{equation}
  r_t=-\lVert n_t-\hat c_t\rVert^2,
  \label{eq:reward}
\end{equation}
\endgroup
on those axes; the undiscounted return sums it over the ensemble and replayed attempts, with a penalty on correction changes~\cite{liao2026supermimic}.
The reach itself has no error term: corrections during the reach count only through the hold accuracy they lead to.
This kinematic reward needs neither task-success labels nor ground-truth object state.

\paragraph{Goal augmentation}
Goal augmentation~\cite{andrychowicz2017her} reuses each attempt for other targets.
Moving $p^\star$ by $d$ shifts $n_t$ and the harness command by $\alpha_t d$ (\cref{eq:contact_alignment}); since the model reads the harness command and the correction only through their sum, shifting the correction by $-\alpha_t d$ leaves the issued commands unchanged, so the attempt is also an exact sample for $p^\star+d$.
Each replay draws $d$ as another attempt's $p^\star$ minus its own, both relative to the shoulder.

\paragraph{Online execution and updates}
The model and actor start from scratch for each task step in each session; during the first \rlWarmup\ attempts the robot explores around zero correction.
Both train in the background after every attempt, and each attempt loads the newest actor that passes a validity check.
After closure the correction ramps to zero, and the frozen tracker executes the rest of the motion, including sustained contact.

\subsection{Slow system: refinement through agent review}
\label{sec:slow}
The slow system fixes failures that a command correction cannot: perception misses the target or segments the wrong object, or the progress indicator accepts an empty grasp.

\paragraph{Proposing revisions}
Outside the control loop, the agent reviews the instruction, saved camera frames, and attempt summaries.
When the phrase is vague or wrong, or the instruction covers several objects, it proposes one specific phrase per object, grounded in turn as the robot lifts each (\cref{fig:slow_real}).
When carry frames show an empty grasp that was judged successful, it revises how the indicator checks the end state.

\paragraph{Validating and applying revisions}
A revision is kept only if it does better when replayed on saved observations: a phrase must raise object detections on held-out frames, net of wrong-object detections, and an indicator revision must reduce disagreement with hindsight labels from later frames without adding false successes.
Accepted revisions take effect at the next attempt.

%% file: arxiv/sec/4_experiments.tex
\begin{figure}[t]
  \centering
  \includegraphics[width=\linewidth]{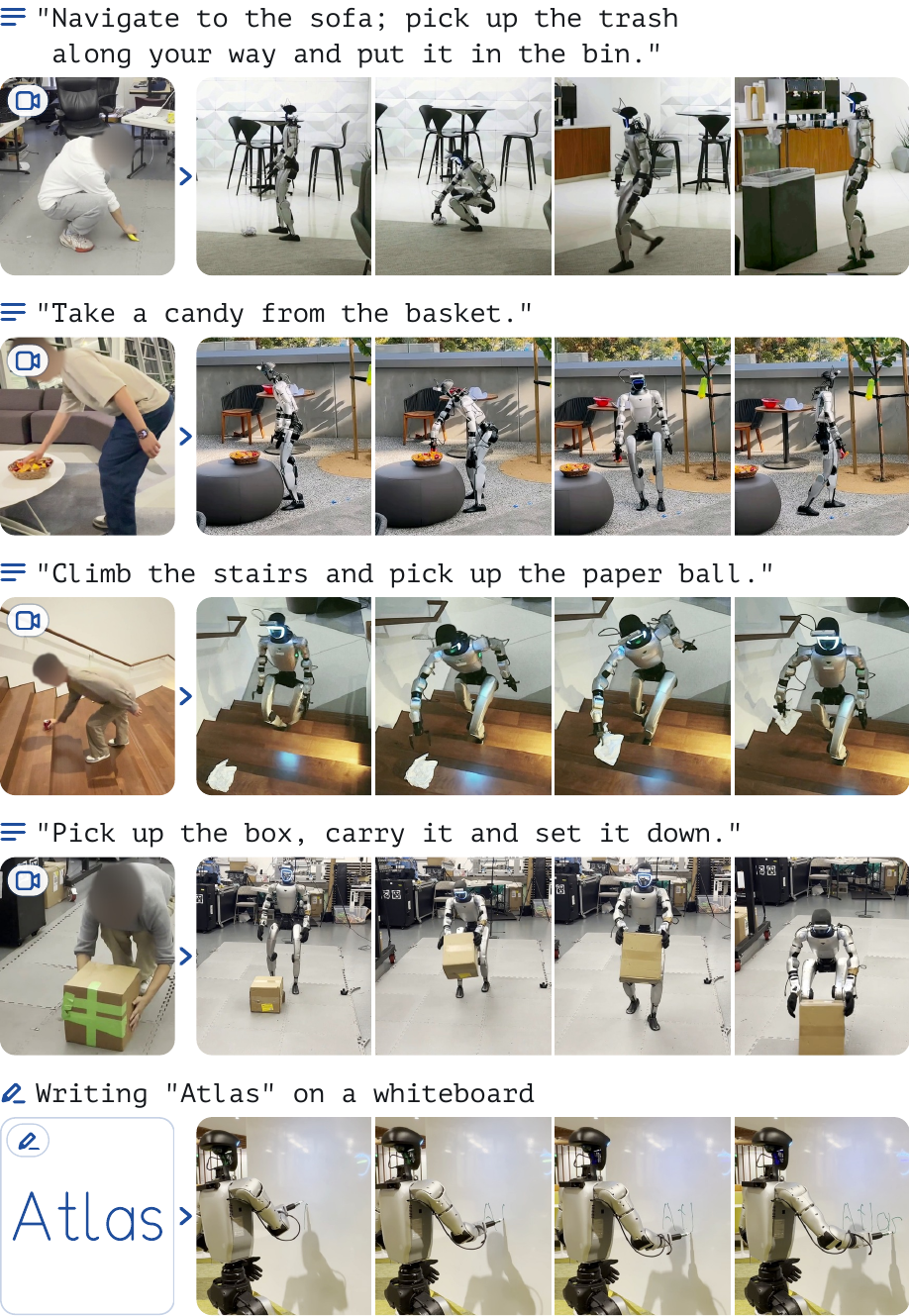}
  \caption{\textbf{Tasks on the real robot.}
  The first four rows each show the prompt, the human video demonstration (left) and the robot's run.
  In the last row the humanoid writes ``Atlas'' on a whiteboard, from the word's strokes (\cref{sec:exp_hw}).}
  \label{fig:tasks}
\end{figure}
\input{arxiv/tab/benchmarks}

\begin{figure*}[t]
  \centering
  \includegraphics[width=\linewidth]{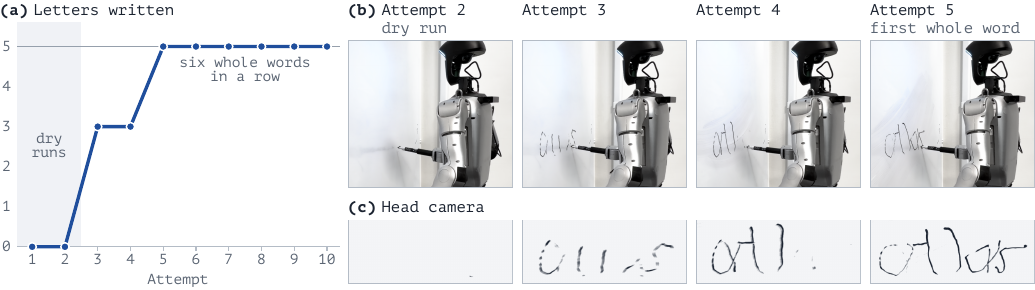}
  \caption{\textbf{The humanoid writes the whole word \wrWholeCountWord\ times in a row.}
  (a)~Letters written in each attempt of the take, out of \wrLettersWord; a letter counts as written when ink lies along more than \wrLetterRule\ of its stroke length in the head camera's images of the board.
  (b)~Attempts~\wrSnapFirst--\wrSnapLast\ at the end of the last stroke, from the last dry run to the first whole word; earlier marks are toned down.
  (c)~The same attempts in the robot's head camera, after each attempt with the pen up; the board's shading is evened out.}
  \label{fig:writing}
\end{figure*}

\section{Experiments and Results}
\label{sec:experiments}
We evaluate whole-body skill acquisition from one human video and one instruction (\cref{sec:exp_harness}) and on diverse hardware tasks (\cref{sec:exp_hw}), the slow system (\cref{sec:exp_slow}), the reuse of motion references across objects, scenes, and object heights (\cref{sec:exp_more}), and the contribution from the fast system  (\cref{sec:ablation}).

\subsection{Setup}
\label{sec:setup}
\paragraph{Robot and simulation}
Real-robot experiments use a Unitree G1 humanoid with onboard RGB-D sensing and joint-state measurements.
The frozen whole-body tracker and actor run on the robot, while an external PC performs vision-language inference, segmentation, and updates to the dynamics model and actor.
We use Claude Opus 5.5 as the agent.
Simulation experiments use Isaac Lab with the same learning and execution framework, and the SIMPLE benchmark.

\subsection{Acquire whole-body skill from human video}
\label{sec:exp_harness}
\paragraph{Human motion supports diverse whole-body interactions}
On the real robot, \method performs object pickup, box transport, and stair traversal with manipulation, each from one related human video and one instruction (\cref{fig:tasks}).

\paragraph{One reference supports competitive performance across three tasks}
\label{sec:sim}
With one human video and no task-specific robot demonstrations, \method succeeds in \simXmpAtlas/30, \simXmbpAtlas/30, and \simBpAtlas/30 episodes on \emph{XMovePick}, \emph{XMoveBendPick}, and \emph{BendPick} of the public SIMPLE loco-manipulation benchmark~\cite{wei2026simple}, within one success of the best VLA, world-action-model, or imitation policy fine-tuned on 100 robot demonstrations per task (\cref{tab:benchmarks}).
For a fair comparison, \method runs in SIMPLE's own environment, with its physics, head camera, reset states, and success check, in the same setting as our other experiments.
Its fast-system learners start from zero at the first episode; because box pickups tolerate centimeter-level end-effector errors, the edited reference already reaches suitable grasp points before any correction is learned, so these scores mainly reflect the harness.

\subsection{Quantitative investigation on diverse hardware tasks}
\label{sec:exp_hw}
\begin{figure*}[t]
  \centering
  \includegraphics[width=\linewidth]{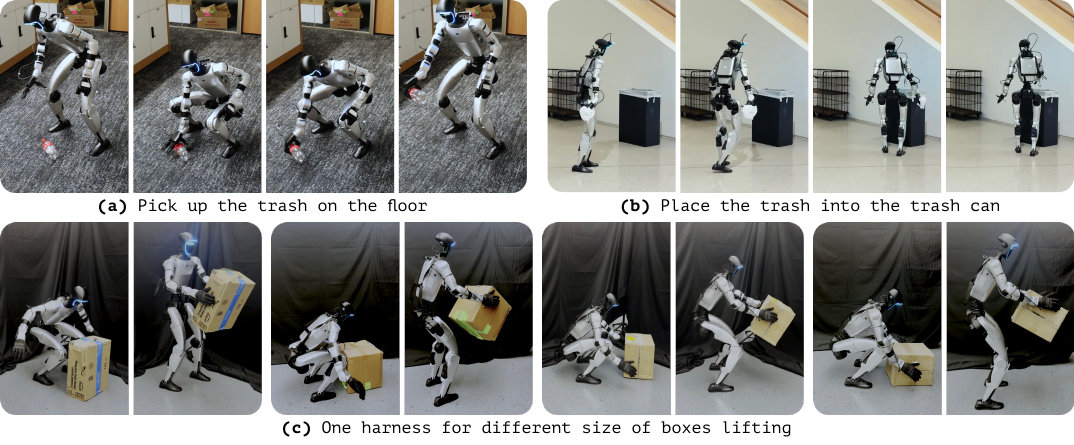}
  \caption{\textbf{The full system completes diverse real-robot tasks.}
  The humanoid (a)~picks trash off the floor, (b)~puts it into a trash can, and (c)~crouches to lift boxes of three sizes with one harness, the smallest both standing and lying.
  Success rates are reported in \cref{tab:real_tasks}.}
  \label{fig:real_tasks}
\end{figure*}

\begin{table}[t]
  \centering
  \footnotesize
  \begin{tabular*}{\linewidth}{@{\extracolsep{\fill}}lc@{}}
    \toprule
    Task & Success \\
    \midrule
    (a) Pick up the trash on the floor & \realPickSucc \\
    (b) Place the trash into the trash can & \realBinSucc \\
    (c) Lift the box from the floor & \realBoxSucc \\
    \bottomrule
  \end{tabular*}
  \caption{\textbf{Success on the real-robot tasks of \cref{fig:real_tasks}.} (a)~The attempts of our learner on the three objects of \cref{tab:real_ablation}. (b)~Trials that ended with the trash in the trash can, out of those in which the robot picked it up. (c)~All three box sizes together.}
  \label{tab:real_tasks}
\end{table}

\paragraph{Writing on a whiteboard}
With a writing harness and the fast system of \method, the humanoid writes the whole word ``atlas'' \wrWholeCountWord\ times in a row on a wall-mounted whiteboard, with a marker fixed to its right wrist (\cref{fig:writing}).
In one uninterrupted \wrTakeWord-minute take, it makes \wrAttemptsWord\ consecutive attempts: \wrProbeWord\ dry runs aimed short of the board, \wrPartialWord\ partial words, and then \wrWholeCountWord\ whole words, with ink along more than \wrCovLateFloor\% of their stroke length on average (\cref{fig:writing}a).
The marker leaves ink only while its tip touches the board, so the harness measures the board's plane with the head camera \wrLiveWord\ times per second and keeps the strokes on it while the standing body moves by up to \wrBoardMoveMm\,mm (\cref{sec:suppl_writing}).

\paragraph{Putting trash into a can and lifting boxes}
The full system picks objects off the floor in \realAllMbK\ of \realAllN\ attempts (\cref{sec:ablation}), puts the trash it picked up into a trash can in all \realBinN\ trials, and lifts boxes of three sizes in \realBoxK\ of \realBoxN\ attempts (\cref{fig:real_tasks,tab:real_tasks}).

\subsection{The slow system improves grounding and success assessment}
\label{sec:exp_slow}
\begin{figure*}[t]
  \centering
  \includegraphics[width=\textwidth]{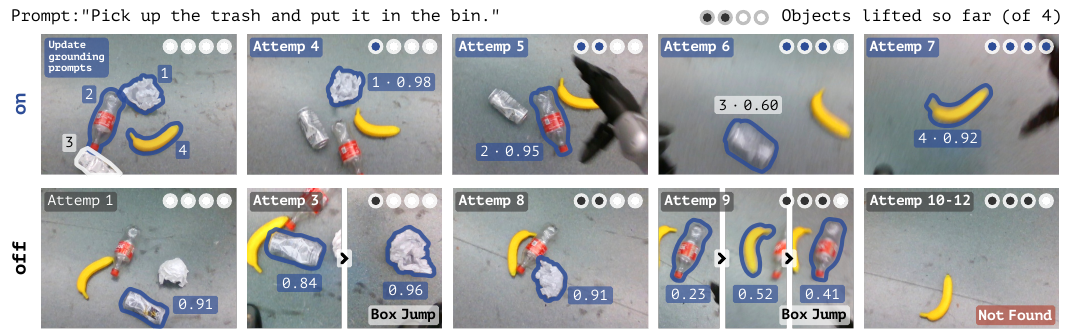}
  \caption{\textbf{The slow system replaces the vague phrase ``the trash.'' with one phrase per object, written by the agent from the robot's own views.}
  Same scene and prompt in both runs; tabs give the attempt and the time since the first look, filled dots the objects lifted so far, and box labels the phrase number and the \segmenterName\ score.
  Top: the agent writes one phrase per object; phrase~3 lands on object~1 and is dropped, and the can is found with phrase~2 once the bottle is gone.
  Bottom: every attempt asks for ``the trash.'', the box jumps between objects, and the banana is never found.}
  \label{fig:slow_real}
\end{figure*}
\paragraph{Online review enables recovery from grounding failures}
Agent review recovers a pickup that starts from a wrong object description: a simulated red-apple pickup given a green-apple description succeeds in \slowLoopOnK/\slowLoopN{} attempts with the slow system and \slowLoopOffK/\slowLoopN{} without it, including failed attempts.
Each condition runs \slowLoopSeedsWord\ sessions of \slowLoopAttemptsPerSeed\ attempts with the fast system enabled.

\paragraph{Revisions improve perception and success assessment}
Revised phrases raise recall on held-out frames of real-robot runs from \slowRecallOffD\% to \slowRecallOnD\%, above a segmenter threshold tuned on the same early frames (\slowRecallThrD\%, Holm $p=\slowRecallThrHolmP$).
The candidates are proposed and selected on early frames of \slowRunsWord\ runs where the initial phrase failed.
Revised success checks reduce disagreements with hindsight labels from \slowIndDisagreeOff\ to \slowIndDisagreeOn\ across \slowIndN\ held-out real-robot attempts: false successes on empty grasps fall from \slowIndFalseSuccOff\% to \slowIndFalseSuccOn\%, while false misses on held objects rise from \slowIndFalseMissOff\% to \slowIndFalseMissOn\%.

\paragraph{The robot collects multiple objects from one instruction}
With the slow system, the robot clears all four objects of a scene from the instruction to pick up the trash, in \slowOnAttempts\ attempts and \slowOnTime; without it, the robot lifts \slowOffLifted\ objects in \slowOffAttempts\ attempts and never finds the banana (\cref{fig:slow_real}).
Both runs start from the phrase ``the trash.''.

\subsection{Reuse motion references across objects, scenes, and heights}
\label{sec:exp_more}
\begin{figure*}[t]
  \centering
  \begin{minipage}[b]{0.3273\textwidth}\centering
    \includegraphics{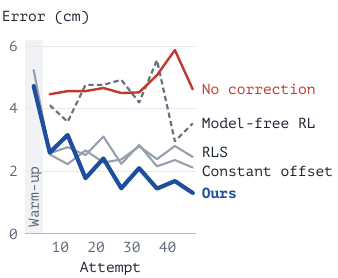}\\[1pt]
    {\footnotesize (a) Error at one placement}
  \end{minipage}%
  \begin{minipage}[b]{0.3163\textwidth}\centering
    \includegraphics{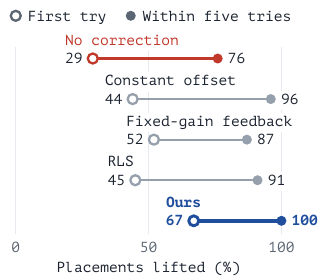}\\[1pt]
    {\footnotesize (b) Retry in place}
  \end{minipage}%
  \begin{minipage}[b]{0.3564\textwidth}\centering
    \includegraphics{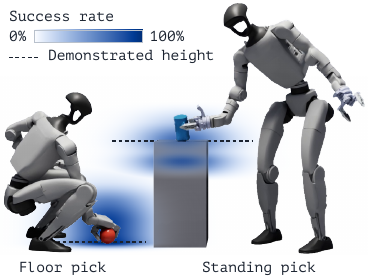}\\[1pt]
    {\footnotesize (c) Reach of one demonstration}
  \end{minipage}
  \caption{\textbf{Our learner keeps lowering the error at one placement, retries recover most first-try misses, and one demonstration covers a band of object heights.}
  (a) Simulated apple pick at one placement: median end-effector error over the hold in bins of five attempts, pooled over three sessions per learner.
  (b) Real robot: share of placements lifted at the first try (hollow) and within five tries (filled), pooled over the objects of \cref{tab:real_ablation}.
  (c) Simulation, with the object raised on a support: each robot is shown at its demonstrated height, and the glow's shade is the success rate over ten attempts at each height, interpolated between  heights.}
  \label{fig:rl}
\end{figure*}
\input{arxiv/tab/real_ablation}
\paragraph{Reuse across objects and scenes}
On the real robot, the same banana-picking reference grasps a banana, a tennis ball, and a paper ball in different scenes (\cref{fig:one_motion}).

\paragraph{Adaptation to object height}
One demonstration covers a band of object heights around the demonstrated one (\cref{fig:rl}c).
The floor-pick reference, demonstrated at 0\,cm, lifts the apple in \hFlZero, \hFlFive, \hFlTen, \hFlTwenty, and \hFlThirty{} of ten attempts at 0, 5, 10, 20, and 30\,cm and never at 40--80\,cm; the standing reference, demonstrated at a \hDemoStandGraspCm\,cm grasp height on a \hDemoStandSupportCm\,cm support, lifts the can in \hStForty, \hStFortyFive, \hStFifty, and \hStSixty{} of ten at 40, 45, 50, and 60\,cm and never at 20, 30, 70, or 80\,cm.
In simulation, we vary support height and reuse the reference and task specification with the height adjustment of \cref{sec:edit}; each height starts with a fresh learner and \rlWarmup\ warm-up attempts before 10 scored trials.

\subsection{The fast system improves execution accuracy and grasp success}
\label{sec:ablation}
\label{sec:exp_inner}

\paragraph{Baselines}
We compare online model-based RL with zero correction and three corrections held fixed within each attempt: a constant offset estimated during warm-up, recursive least squares (RLS) updating a bias and gain after each attempt, and, on the real robot, fixed-gain feedback subtracting the preceding attempt's error.
All share the correction channel, bounds, and \rlWarmup\ warm-up attempts; our actor adjusts the correction at every control step during the reach and hold, with its weights fixed within the attempt.
Simulation also includes a model-free actor-critic acting at every control step, with the same correction increments, bounds, and exploration noise.
        
\paragraph{Model-based RL improves accuracy beyond offset estimation}
Our learner keeps lowering the error after the offset estimators level off, ending at \appMbLast\,cm against \appCoLast\,cm for the constant offset, \appRlsLast\,cm for RLS, and \appZeLast\,cm without a correction (\cref{fig:rl}a).
From attempt 25 on, its median error is \appMbLateMed\,cm against \appCoLateMed\,cm and \appRlsLateMed\,cm, lower than both estimators in each of the three sessions ($p=\appLateCoP$ against the constant offset, two-sided Mann--Whitney U test over the attempts of the three sessions).
The model-free actor-critic, which also changes the command at every control step, ends at \appMfLast\,cm under the same interaction budget, so acting at every step does not by itself bring the gain.
We run three 50-attempt sessions per learner at a fixed simulated apple placement, interleaving zero-correction attempts after warm-up; the error is the median $\lVert c_t-n_t\rVert$ over the pre-closure hold, and the constant offset, RLS, and our learner all reach about \appFirstBin\,cm right after warm-up.

\paragraph{The model learns more than a constant tracking bias}
The learned model predicts held-out trajectories better than a constant bias: open loop from their first measured state, its error falls from \offMbFive\,cm after five training attempts to \offMbForty\,cm after 40, against \offCoForty\,cm for the command plus a fitted constant offset.

\paragraph{Online correction improves pickup success}
On seeded random placements in simulation, our learner raises first-try lift success from \tbFirstZe\% without a correction to \tbFirstMb\%, and success within ten attempts from \tbTenZe\% to \tbTenMb\%.
With the simulator's exact targets and robot poses, the offset estimators reach similar rates.

\paragraph{The fast system achieves higher real-robot success}
On real-robot floor pickups, our learner has the highest success on every object, \realAllMbK/\realAllN{} attempts in total versus \realAllZeK/\realAllN{} without a correction, and the lowest error on the can and apple (\cref{tab:real_ablation}).
Retrying in place then lifts all \rtyAllMbN\ eligible placements within five tries, up from \rtyAllMbF\% at the first try, while the control rises from \rtyAllZeF\% to \rtyAllZeW\% (\cref{fig:rl}b).
We evaluate banana, can, and apple pickups with a three-finger end effector in separate sessions, scoring 20 attempts per method and object after learner warm-up.
Success requires standing up holding the object; tracking error, the distance from the contact point to its target $n_t$ (which the correction does not move), is measured at contact.
Whether a grasp at that target holds also depends on perception and on the object, so error and success need not rank the learners alike.
Contact and target calibrations vary across sessions, and our apple run also uses a camera correction, so these runs compare deployed configurations.

%% file: arxiv/tab/benchmarks.tex
\providecommand{\lv}[1]{\textcolor{simpleLvl}{#1}}
\definecolor{simpleLvl}{gray}{0.55}
\newsavebox{\simpleBenchmarkBox}
\begin{table*}[t]
  \centering
  \begin{lrbox}{\simpleBenchmarkBox}
  \begin{minipage}[c]{0.715\linewidth}
  \footnotesize
  \setlength{\tabcolsep}{2.4pt}
  \renewcommand{\arraystretch}{1.05}
  \setlength{\aboverulesep}{0.3ex}
  \setlength{\belowrulesep}{0.4ex}
  \begin{tabular*}{\linewidth}{@{\extracolsep{\fill}}l c cccc cccc cccc@{}}
    \toprule
    & Training & \multicolumn{4}{c}{XMovePick} & \multicolumn{4}{c}{XMoveBendPick} & \multicolumn{4}{c@{}}{BendPick} \\
    \cmidrule(lr){3-6}\cmidrule(lr){7-10}\cmidrule(l){11-14}
    Method & demos & \lv{L0} & \lv{L1} & \lv{L2} & Total & \lv{L0} & \lv{L1} & \lv{L2} & Total & \lv{L0} & \lv{L1} & \lv{L2} & Total \\
    \midrule
    $\Psi_0$~\cite{psi0_2026} & 100 robot & \lv{10} & \lv{10} & \lv{6}  & \simXmpPsi            & \lv{10} & \lv{9} & \lv{9} & \simXmbpPsi          & \lv{10} & \lv{10} & \lv{10} & \textbf{\simBpPsi} \\
    GR00T N1.6~\cite{gr00tn1} & 100 robot & \lv{10} & \lv{10} & \lv{7}  & \simXmpGroot          & \lv{4}  & \lv{4} & \lv{1} & \simXmbpGroot        & \lv{7}  & \lv{7}  & \lv{6}  & \simBpGroot \\
    $\pi_{0.5}$~\cite{pi05} & 100 robot & \lv{7}  & \lv{5}  & \lv{1}  & \simXmpPi             & \lv{0}  & \lv{0} & \lv{0} & \simXmbpPi           & \lv{10} & \lv{10} & \lv{8}  & \simBpPi \\
    DreamZero~\cite{ye2026dreamzero} & 100 robot & \lv{10} & \lv{10} & \lv{10} & \textbf{\simXmpDz}    & \lv{0}  & \lv{0} & \lv{1} & \simXmbpDz           & \lv{9}  & \lv{9}  & \lv{8}  & \simBpDz \\
    ACT~\cite{zhao2023act} & 100 robot & \lv{10} & \lv{10} & \lv{5}  & \simXmpAct            & \lv{9}  & \lv{10} & \lv{10} & \textbf{\simXmbpAct} & \lv{10} & \lv{9}  & \lv{9}  & \simBpAct \\
    Diffusion Policy~\cite{chi2023diffusion} & 100 robot & \lv{3}  & \lv{3}  & \lv{2}  & \simXmpDp             & \lv{0}  & \lv{0} & \lv{0} & \simXmbpDp           & \lv{10} & \lv{8}  & \lv{6}  & \simBpDp \\
    \midrule
    \textbf{\method{} (ours)} & 1 human & \lv{10} & \lv{10} & \lv{9} & \simXmpAtlas & \lv{9} & \lv{9} & \lv{10} & \simXmbpAtlas & \lv{10} & \lv{10} & \lv{10} & \textbf{\simBpAtlas} \\
    \bottomrule
  \end{tabular*}
  \end{minipage}
  \end{lrbox}%
  \usebox{\simpleBenchmarkBox}\hfill
  \begin{minipage}[c]{0.265\linewidth}
    \centering
    \includegraphics[width=\linewidth,
      height=\dimexpr\ht\simpleBenchmarkBox+\dp\simpleBenchmarkBox\relax,
      keepaspectratio]{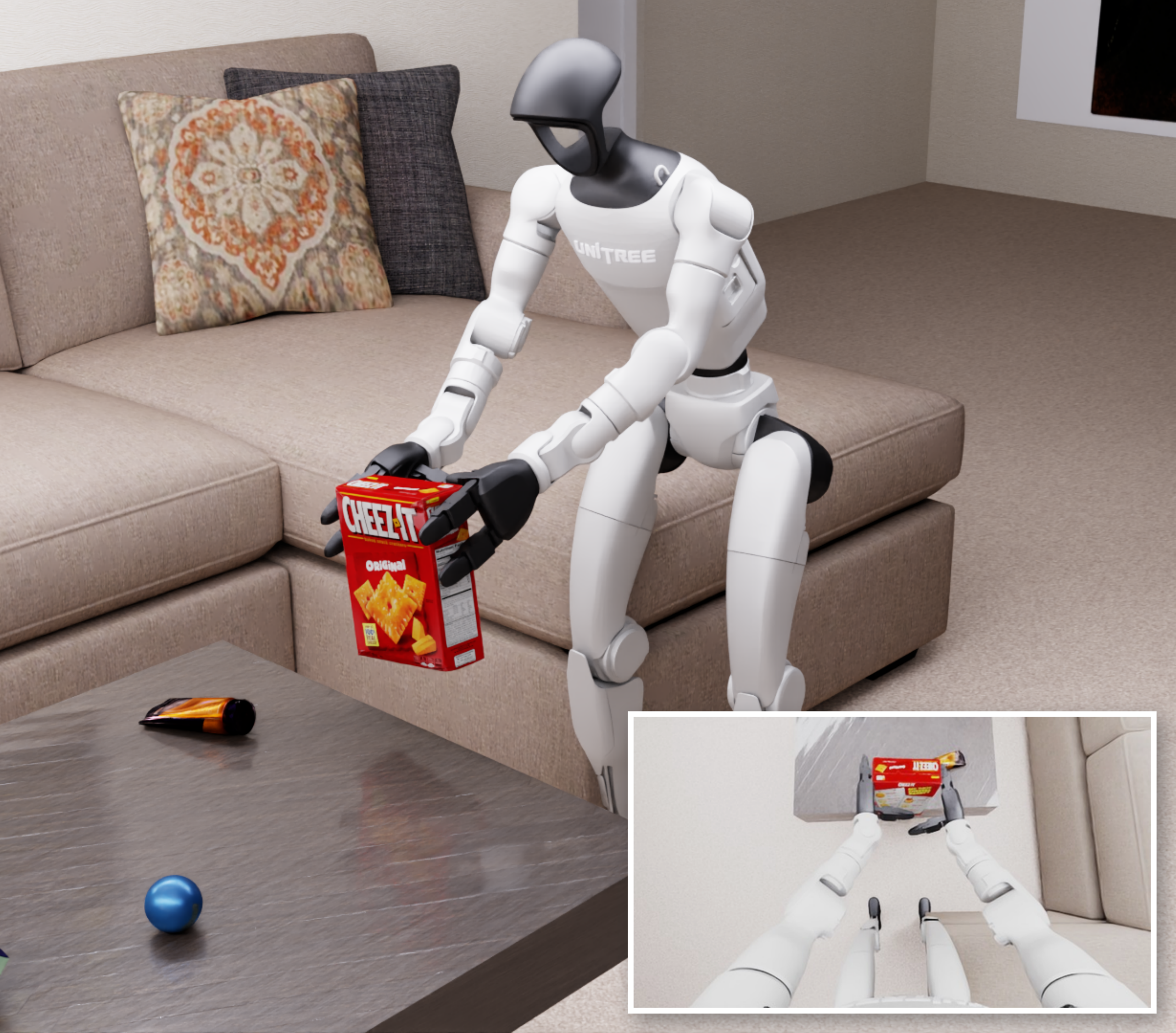}
  \end{minipage}
  \caption{\textbf{On the public SIMPLE benchmark, one human video is competitive with policies fine-tuned on 100 robot demonstrations per task.}
  Successes per level (/10) and overall (/30), using SIMPLE's success check.
  L0: distractors/materials; L1 adds extreme lighting; L2 adds box-pose perturbations.
  Baselines are fine-tuned on 100 robot demonstrations per task; scores from SIMPLE~\cite{wei2026simple}.
  \method{} reuses one human video across all tasks with onboard sensing and no task-specific robot demonstrations.
  Right: SIMPLE render of XMoveBendPick with a simultaneous head-camera inset.
  \textbf{Bold}: best total.}
  \label{tab:benchmarks}
\end{table*}

%% file: arxiv/tab/real_ablation.tex
\begin{table*}[t]
  \centering
  \footnotesize
  \setlength{\tabcolsep}{2pt}
  \renewcommand{\arraystretch}{1.0}
  \setlength{\aboverulesep}{0.3ex}
  \setlength{\belowrulesep}{0.4ex}
  \newcommand{\kn}[2]{{\scriptsize\,(#1/#2)}}
  \newcommand{\cell}[5]{#1\kn{#2}{#3} & #4\ / #5}
  \newcommand{\markbest}[1]{\textbf{#1}}\newcommand{\marksecond}[1]{\underline{#1}}
  \newcommand{\blank}{\multicolumn{2}{c}{}}
  \newcommand{\blanklast}{\multicolumn{2}{c@{}}{}}
  \newcommand{\leafheads}{Success (\%)\up & Error (cm)\down}
  \begin{tabular*}{\textwidth}{@{\extracolsep{\fill}}l cc cc cc@{}}
    \toprule
    \multirow{2}{*}{Learner} & \multicolumn{2}{c}{Banana} & \multicolumn{2}{c}{Can} & \multicolumn{2}{c@{}}{Apple} \\
    \cmidrule(lr){2-3}\cmidrule(lr){4-5}\cmidrule(l){6-7}
    & \leafheads & \leafheads & \leafheads \\
    \midrule
    No correction (control)          & \cell{\realBaZePctD}{\realBaZeK}{\realBaZeN}{\realBaZeCpMean}{\realBaZeCpMed} & \cell{\realCaZePctD}{\realCaZeK}{\realCaZeN}{\realCaZeCpMean}{\realCaZeCpMed} & \cell{\realApZePctD}{\realApZeK}{\realApZeN}{\realApZeCpMean}{\realApZeCpMed} \\
    Constant offset                & \cell{\realBaCoPctD}{\realBaCoK}{\realBaCoN}{\realBaCoCpMean}{\realBaCoCpMed} & \cell{\marksecond{\realCaCoPctD}}{\realCaCoK}{\realCaCoN}{\realCaCoCpMean}{\realCaCoCpMed} & \cell{\marksecond{\realApCoPctD}}{\realApCoK}{\realApCoN}{\realApCoCpMean}{\realApCoCpMed} \\
    Fixed-gain feedback            & \cell{\marksecond{\realBaFbPctD}}{\realBaFbK}{\realBaFbN}{\markbest{\realBaFbCpMean}}{\markbest{\realBaFbCpMed}} & \cell{\realCaFbPctD}{\realCaFbK}{\realCaFbN}{\realCaFbCpMean}{\realCaFbCpMed} & \cell{\realApFbPctD}{\realApFbK}{\realApFbN}{\marksecond{\realApFbCpMean}}{\marksecond{\realApFbCpMed}} \\
    Recursive least squares        & \cell{\realBaRlsPctD}{\realBaRlsK}{\realBaRlsN}{\realBaRlsCpMean}{\realBaRlsCpMed} & \cell{\realCaRlsPctD}{\realCaRlsK}{\realCaRlsN}{\marksecond{\realCaRlsCpMean}}{\marksecond{\realCaRlsCpMed}} & \cell{\realApRlsPctD}{\realApRlsK}{\realApRlsN}{\realApRlsCpMean}{\realApRlsCpMed} \\
    \textbf{Online model-based RL (ours)} & \cell{\markbest{\realBaMbPctD}}{\realBaMbK}{\realBaMbN}{\marksecond{\realBaMbCpMean}}{\marksecond{\realBaMbCpMed}} & \cell{\markbest{\realCaMbPctD}}{\realCaMbK}{\realCaMbN}{\markbest{\realCaMbCpMean}}{\markbest{\realCaMbCpMed}} & \cell{\markbest{\realApMbPctD}}{\realApMbK}{\realApMbN}{\markbest{\realApMbCpMean}}{\markbest{\realApMbCpMed}} \\
    \bottomrule
  \end{tabular*}
  \caption{\textbf{Our learner achieves the highest observed success on all three real-robot pickups.}
  The operator freely places floor objects in front of the robot, which picks with its right end effector and retries misses in place.
  We score 20 attempts per method and object: the control's first 20; learners' first 20 after \rlWarmup\ warm-up attempts, except can fixed-gain feedback (last 20).
  Success requires standing up holding the object; parentheses give successes/attempts.
  Error: mean / median distance from the end-effector contact point to its target at contact.
  Sessions differ in contact and target calibration; our apple session also uses a camera correction (\cref{sec:ablation}).
  Bold: best among learners; underlined: second.}
  \label{tab:real_ablation}
\end{table*}

%% file: arxiv/sec/5_conclusion.tex
\section{Conclusion}
\label{sec:conclusion}
\method is an adaptive harness between a pretrained vision-language agent and a frozen whole-body tracker.
From one related human video and an instruction, it adapts the motion reference and learns from robot attempts to refine grounding and correct end-effector commands.
Real-robot deployments demonstrate interactions from object pickup to stair traversal with manipulation, and reference reuse across objects and scenes.
On SIMPLE, one human video yields \simXmpAtlas/30, \simXmbpAtlas/30, and \simBpAtlas/30 successes, within one success of the best listed baseline on each task; the baselines use 100 robot demonstrations per task for fine-tuning.
In fixed-placement simulation, online model-based RL ends at \appMbLast\,cm median tracking error versus \appMfLast\,cm for model-free RL under the same interaction budget.
Across real-robot configurations with differing calibrations, our learner achieves the highest observed pickup success on all three objects.
Agent review recovers simulated pickup from an incorrect object description; held-out replay of robot observations shows recall rising from \slowRecallOffD\% to \slowRecallOnD\% and fewer false success judgments.
These behaviors use no task-specific robot demonstrations or offline task-policy training, and all pretrained models stay frozen during deployment.

\paragraph{Limitations}
The learned dynamics model is specific to one task step and session, so it does not yet scale across tasks.
Grounding from phrases and depth is also not yet precise enough for fine-grained interactions, which the pipeline cannot yet perform.

\def\mainTextEndLog{\write-1{MAIN-TEXT-END p.\thepage, \the\budgetowed\space lines owed by the stubs}}
\ifbudgetdraft\par\noindent{\scriptsize\color{red}MAIN-TEXT-END}\mainTextEndLog\par\else\par\mainTextEndLog\fi
\ifbudgetdraft\ifnum\value{page}>\mainTextPages\relax\PackageError{budget}{Main text ends on p.\thepage}{Cut before adding text}\fi\fi

%% file: arxiv/sec/6_ack.tex
\section{Acknowledgement}

This work was carried out at UC Berkeley in academic collaboration with the Autonomous System \& Physical AI Research Group (ASPIRE) at NVIDIA. We thank our colleagues and lab members for helpful discussions and support throughout the project. Pieter Abbeel holds concurrent appointments as a professor at UC Berkeley and as an Amazon Scholar; this paper describes work performed at UC Berkeley and is not associated with Amazon.

%% file: arxiv/sec/appendix.tex
\nocite{nvidia2026isaaclabarena,nvidia2026arenastatic}
This appendix first details the whiteboard writing of \cref{sec:exp_hw} and describes how a force loop sets the contact depth in force-involved tasks, with writing in simulation as the example (\cref{sec:suppl_writing}), and then reports further experiments on each part of \method.
\Cref{sec:suppl_fast_results} ablates the fast system's learner one component at a time and tests it on a second task.
\Cref{sec:suppl_harness_results} compares the harness with direct replay of the human motion and runs it on \ctlTrackersWord\ whole-body trackers.
It also compares harnesses written by \agtAgentsWord\ different agents.
\Cref{sec:suppl_slow_results} evaluates the slow system on real-robot frames and online in simulation.
\Cref{sec:suppl_spec,sec:suppl_learner,sec:suppl_agent} give the method details, and \cref{sec:suppl_sim,sec:suppl_protocol} the simulated and real-robot protocols.

\input{arxiv/sec/suppl/results_writing}

\input{arxiv/sec/suppl/results_fast}
\input{arxiv/sec/suppl/results_harness}

\input{arxiv/sec/suppl/results_slow}
\input{arxiv/sec/suppl/spec}
\input{arxiv/sec/suppl/learner}
\input{arxiv/sec/suppl/agent}
\input{arxiv/sec/suppl/sim}
\input{arxiv/sec/suppl/protocol}

%% file: arxiv/sec/suppl/results_writing.tex
\section{Writing on a whiteboard}
\label{sec:suppl_writing}

\paragraph{Writing needs the marker on the board through every stroke, so the harness follows the board as the body moves}
The marker leaves ink only while its tip touches the board, the word takes \wrStrokeM\,m of strokes, and the standing body moves by up to \wrBoardMoveMm\,mm relative to the board within a single word.
The harness measures the board's plane from the depth image of the robot's own head camera before every attempt and \wrLiveWord\ times per second while writing, and places the strokes' path on that plane.
The reference comes from the word's strokes: it holds a standing pose while the right wrist traces the \wrWordWidthCm\,cm-wide word.
The fast system acts within the board's plane and treats each letter as a separate attempt.
From the first attempt on the board to the last, no one touches the robot: a person only starts each attempt and wipes the board.
\Cref{fig:tasks} (last row) shows a later session, in which it writes ``Atlas''.

\begin{figure*}[t]
  \centering
  \includegraphics[width=\textwidth]{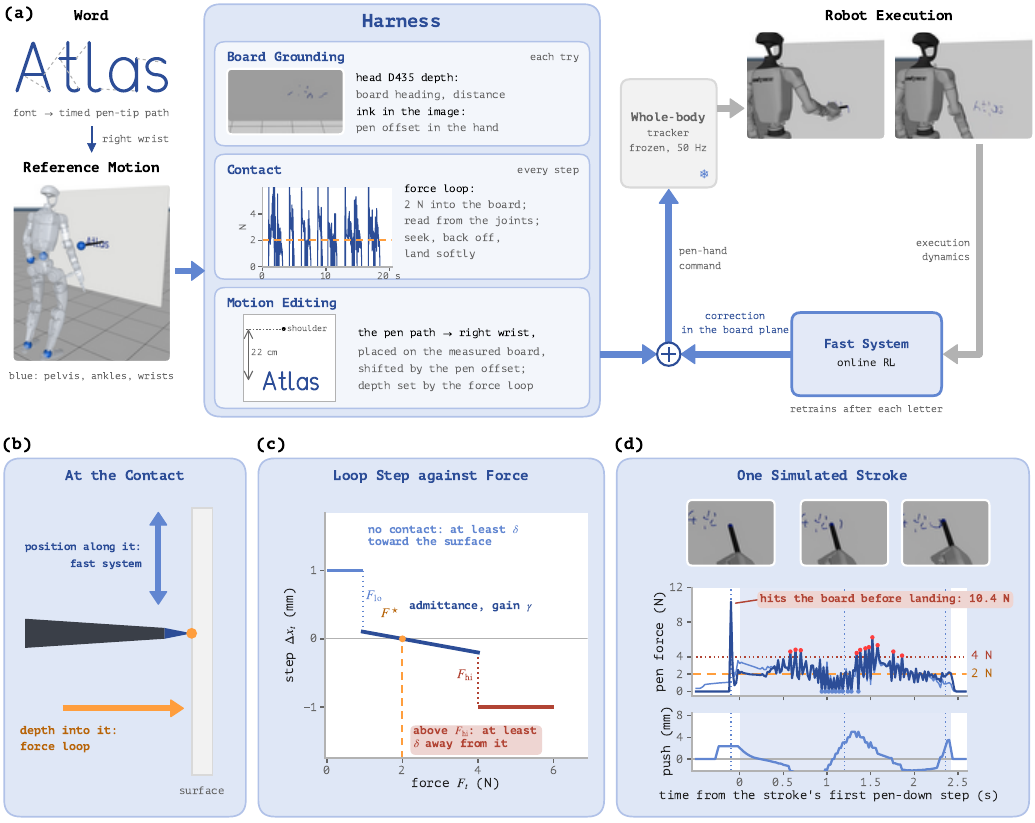}
  \caption{\textbf{A force loop sets the depth into the surface, and the fast system the position along it.}
  (a)~The writing pipeline, in simulation: the word becomes a timed pen-tip path that the right wrist carries in a reference motion; the harness grounds the board from the head camera, runs the force loop on the force estimated from the arm's joints, and places the path on the measured board; the fast system adds its correction in the board plane to the pen-hand command, which the frozen tracker executes.
  The pictures are the simulation's: the reference motion, the head camera's view of the board, and two frames of the robot writing (from a re-run with the loop on the simulator's force).
  (b)~At the contact, the depth into the surface is the force loop's and the position along it the fast system's.
  (c)~The loop's step $\Delta x_t$ against the force once the landing ramp is over ($\lambda_t=1$), with the values of the writing runs: an admittance of gain $\gamma$ around the target force $F^\star$, and a step of at least $\delta$ below $F_{\mathrm{lo}}$, the contact threshold of the joint estimate, and above $F_{\mathrm{hi}}$.
  (d)~One typical stroke of the simulated writing with the loop on the joint estimate, the letter ``a'' of try 6, with the simulation's head-camera frames at the dotted lines: the pen's force from the simulator's sensor (dark) and its estimate from the joints, which the loop reads (blue), with dots where the sensor reads no contact with the pen down (blue) or more than $F_{\mathrm{hi}}$ (red), and the push sent to the tracker.}
  \label{fig:force_loop}
\end{figure*}

\paragraph{When the end effector presses on a surface, a force loop and the fast system can together form a contact layer}
Writing is one of several tasks in which the end effector presses on a surface, as in wiping or pressing a button.
At the contact the command then has two directions (\cref{fig:force_loop}b).
Along the surface, the error is a position, which the fast system corrects as for the pickups.
Into the surface, the quantity to control is the force, which the commanded depth sets through the arm's compliance: the surface stops the end effector, and the arm's stiffness turns a depth past the surface into a pressure.
We give the depth to a force loop that reads the force; the loop and the fast system together form a contact layer of the harness, and the frozen tracker is unchanged.
With $\nu$ the unit normal into the surface, from perception (for the board, the plane measured above), the target sent to the tracker is
\begin{equation}
  n_t+u_t+\bigl(\bar x_{k(t)}+w_t\,x_t\bigr)\,\nu,
  \label{eq:contact_command}
\end{equation}
where $n_t$ is the planned path, which lies a small press past the surface, $u_t$ is the fast system's correction, acting along the surface, $k(t)$ is the contact segment of the reference at step $t$, $\bar x_k$ is that segment's depth prior, $x_t$ is the loop's push and $w_t$ is the weight that fades it when the end effector is off the surface.

\paragraph{The force loop moves the push toward a target force, with switches for contact and back-off}
On each step of planned contact, the loop reads the force $F_t$ along the normal and moves the push $x_t$, positive into the surface and zero at the start of each attempt, toward a target force $F^\star$ (\cref{fig:force_loop}c):
\begin{equation}
  \begin{gathered}
  x_t=\operatorname{clip}\bigl(x_{t-1}+\Delta x_t\bigr),\\
  \Delta x_t=\begin{cases}
    \max\{\gamma(\lambda_tF^\star-F_t),\,\delta\} & F_t\le F_{\mathrm{lo}},\\
    \min\{\gamma(\lambda_tF^\star-F_t),\,-\delta\} & F_t> F_{\mathrm{hi}},\\
    \gamma(\lambda_tF^\star-F_t) & \text{otherwise.}
  \end{cases}
  \end{gathered}
  \label{eq:force_loop}
\end{equation}
The last line is an admittance with gain $\gamma$.
Two switches make the loop faster than the admittance alone: without contact ($F_t\le F_{\mathrm{lo}}$) the push advances by at least $\delta$ per step toward the surface, and above $F_{\mathrm{hi}}$ it backs off by at least as much.
The target ramps in over the first $J$ steps of each segment, $\lambda_t=\min(1,j/J)$ on the $j$-th step counted from the landing, so that the loop does not read the low force of the landing as missing push.
The push is clipped to a fixed range and held, not updated, between segments.
What is sent is $x_t$ times $w_t$, which is 1 in contact, falls to 0 within a few steps after the end effector lifts, so that the push does not hold it on the surface, and is whole again a few steps before the next landing.

\paragraph{A depth prior carries the push of one attempt to the next}
The arm reaches different segments in different postures, so one depth cannot serve them all.
After each attempt, each segment's prior moves by the push the loop needed on it,
\begin{equation}
  \bar x_k\leftarrow\operatorname{clip}\bigl(\bar x_k+\eta\,\operatorname{median}_{t\in k}x_t\bigr),
  \label{eq:depth_prior}
\end{equation}
over the segment's steps in contact, within a fixed range, and unchanged after a fall.
So the depth is learned online across attempts, while the fast system learns the position along the surface with online RL: each attempt starts near the depth each segment needs, and the loop handles the variation within a segment.

\paragraph{We ran the contact layer on writing in simulation, with the following values}
In simulation (mjlab), the humanoid writes a word with a pen on a whiteboard (\cref{fig:force_loop}a).
The word is given; it becomes a timed path of the pen tip from a single-stroke rounded font, and the right wrist's target carries that path while the robot stands.
The harness grounds the board as above, which gives $\nu$, and the robot turns or steps to face the board when it is off.
The segments are the letters, each one attempt of the fast system (\cref{sec:suppl_learner}); after two warm-up letters each letter is dealt the zero, mean or explore mode with equal probability, so the letters written without correction form a control within the session.
The layer's values are the following.
The target force is $F^\star=\wrForceN$\,N, a light marker pressure, and the plan lies \wrPressMm\,mm past the board's face.
The loop runs at \controlRateHz\,Hz with gain $\gamma=\wrGainMm$\,mm per newton of error per step; the switches act below $F_{\mathrm{lo}}=\wrSeekBelowJointsN$\,N, the contact threshold of the joint estimate (\wrSeekBelowN\,N on a force sensor), and above $F_{\mathrm{hi}}=\wrBackoffAboveN$\,N, twice the target, by $\delta=\wrSeekMm$\,mm per step; the target ramps in over $J=\wrRampSteps$ steps; and the push stays within $-\wrBoundLoCm$ and $+\wrBoundHiCm$\,cm.
The weight $w_t$ falls to 0 within \wrFadeSteps\ steps after a lift and is whole again \wrFadeLeadSteps\ steps before the next landing.
The prior moves at $\eta=\wrDepthRate$ within $\pm\wrDepthMaxCm$\,cm.

\paragraph{The loop reads a force estimated from the arm's joints as well as it reads a sensor, and an actor holding the pressure keeps the pen on the board far less}
The loop needs no force sensor: it reads the board's push on the pen from the right arm's joints.
From the joint targets and the realized joint state at \controlRateHz\,Hz, it takes the torque the arm's PD controller applied, subtracts what the arm's dynamics need without the push, and solves for the push along the arm's seven joints (\cref{fig:force_loop}a).
The estimate differs from the simulator's contact force by about \wrJointsRmsN\,N RMS, so its contact threshold is higher, \wrSeekBelowJointsN\,N.
The simulator's force, the board's push on the pen averaged over the control step's four physics steps, is the ground truth of what follows, and the loop can read it as well.
A session is 12 tries of the word ``Atlas'' with 50\,mm letters, on one seed; the pen is on the board when the simulator's force is above \wrSeekBelowN\,N, and the error of a letter is the median distance in the board's plane between the tip and its target at the same step, over the letters of tries 6--11 written with the learner's mean correction.
With the loop on the joint estimate, the pen is on the board for \wrBoardJoints\% of the pen-down steps of the session and the error is \wrErrJoints\,mm; with the loop on the simulator's force, \wrBoardLoop\% and \wrErrLoop\,mm.
Over the same tries, the letters written with no correction, the control, err by \wrErrJointsZero\ and \wrErrLoopZero\,mm.
We also let the fast system's actor hold the pressure itself, with the loop and the depth prior off: the actor then acts on the depth axis too and reads the pen's force, with one network or with two for the model.
It keeps the pen on the board for only \wrBoardActorOne\% and \wrBoardActorTwo\% of the pen-down steps, with errors of \wrErrActorOne\ and \wrErrActorTwo\,mm, so we use the loop for the depth and the fast system for the position along the surface.

%% file: arxiv/sec/suppl/results_fast.tex
\section{Additional experiments: the fast system}
\label{sec:suppl_fast_results}
Every experiment here runs the fast system (\cref{sec:fast,sec:suppl_learner}) in Isaac Lab (\cref{sec:suppl_sim}).
The last two expand on the results of \cref{sec:ablation}.
An attempt's hold error is the median $\lVert c_t-n_t\rVert$ over its pre-closure hold, as in \cref{fig:rl}a.

\begin{figure*}[!t]
  \centering
  \includegraphics{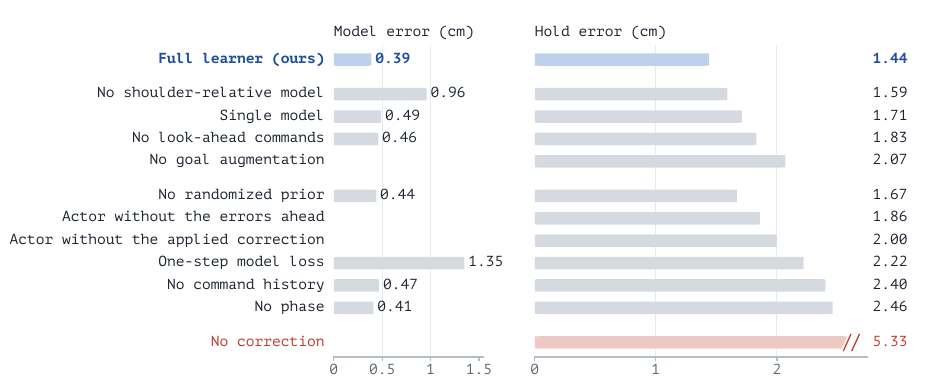}
  \caption{\textbf{The full learner has the lowest hold error.}
  Floor pick at one placement, in new sessions under the protocol of \cref{fig:rl}a.
  Hold error: median over a session's attempts in which the actor acts without exploration, averaged over three sessions; the last six variants ran one session.
  Model error: held-out prediction error after 40 training attempts (\cref{tab:suppl_model_check}), for the variants that change the model.
  No correction: the attempts with no correction in the full learner's sessions.}
  \label{fig:suppl_ablation}
\end{figure*}

\paragraph{To test whether each component of the learner is needed, we remove one at a time}
Removing any one component of the learner raises its hold error (\cref{fig:suppl_ablation}).
We repeat the floor-pick sessions of \cref{fig:rl}a without each component, rerunning the full learner alongside.
All sessions use the same placement and warm-up.
Sessions at the same seed give each attempt the same mode: no correction, the actor, or the actor exploring (\cref{sec:suppl_sim}).
\begin{itemize}[leftmargin=*,itemsep=1pt,topsep=2pt]
  \item \emph{No shoulder-relative model.} The model's inputs and output, and the actor's copies of these inputs, are taken relative to the floor point under the pelvis at the attempt's start instead of the moving shoulder $b_t$. The model must then also predict the body's motion.
  \item \emph{Single model.} One member replaces the ensemble of \rlMembers\ members.
  \item \emph{No look-ahead commands.} The model's input keeps the current command and drops the look-ahead commands after it.
  \item \emph{No goal augmentation.} The actor is trained on each replayed attempt with its own target only, without shifted targets.
  \item \emph{No randomized prior.} The \rlMembers\ members lose their fixed random prior and differ only in their initialization.
  \item \emph{Actor without the errors ahead.} The actor's state keeps the current error $n_t-c_t$ and drops the errors \rlNeedLo\ and \rlNeedHi\ steps ahead.
  \item \emph{Actor without the applied correction.} The actor's state drops the applied correction and keeps the correction in flight.
  \item \emph{One-step model loss.} The model is fit with a one-step prediction loss from each measured state instead of the multi-step loss of \cref{sec:fast}.
  \item \emph{No command history.} The model's input drops the \rlHistPrev\ previous commands.
  \item \emph{No phase.} The model's input drops the reference phase.
\end{itemize}
The actor still reads every input dropped from the model.

The full learner's hold error averages \ablFullM\,cm over three sessions, and a single model is worse in all three, at \ablOneM\,cm on average.
Without the shoulder-relative frame, the model's held-out prediction error more than doubles, from \ablOffFull\ to \ablOffNoSh\,cm, and the mean hold error rises to \ablNoShM\,cm.
Without the look-ahead commands, it rises to \ablNoLookM\,cm, and without goal augmentation to \ablNoGoalM\,cm.
Each of the six further variants, run once, also stays above the full learner's \ablFullM\,cm.
Without the reference phase, the error rises to \ablNoPhase\,cm, and without the command history to \ablNoHist\,cm.

\paragraph{To test whether the learner carries over to another task, we run it on a standing apple pick}
The learner cuts the hold error from \ablStandZero\ to \ablStandOurs\,cm (\cref{tab:suppl_two_tasks}).
The task reuses the standing reference of \cref{fig:rl}c with the floor pick's apple and three-finger end effector.
The apple rests on a 50\,cm support.
Every one of the \ablStandN\ attempts in which the actor acts without exploration stays within \ablStandHi\,cm, including the first.

\begin{table}[!htbp]
  \centering
  \footnotesize
  \begin{tabular}{@{}l cc@{}}
    \toprule
    & \multicolumn{2}{c@{}}{Hold error (cm)\down} \\
    \cmidrule(l){2-3}
    Task & No correction & Ours \\
    \midrule
    Floor pick (three sessions)                 & \ablZeroM     & \textbf{\ablFullM} \\
    Standing apple pick (one session)           & \ablStandZero & \textbf{\ablStandOurs} \\
    \bottomrule
  \end{tabular}
  \caption{\textbf{The learner cuts the hold error on both tasks.}
  Median hold error over the attempts with no correction and over those in which the actor acts without exploration, within the same sessions.
  Floor pick: the means of \cref{fig:suppl_ablation}.
  On the standing apple pick, the robot localizes from the simulator's root pose, as in \cref{fig:rl}c.}
  \label{tab:suppl_two_tasks}
\end{table}

\paragraph{To test whether the dynamics model learns more than a constant offset, we compare their predictions}
After 40 training attempts, the learned model's open-loop prediction is off by \offMbForty\,cm on average, a tenth of the error of a constant offset (\cref{tab:suppl_model_check}).
We refit the model offline with the learner's own code and schedule.
The data are the apple sessions of \cref{fig:rl}a with our learner, the constant offset and recursive least squares, nine in all.
For each session, the model trains on its first $k$ attempts and is tested on its attempts after the 40th.
From a test attempt's first measured state, it predicts the end effector's position over the whole reach and hold.
It reads only the recorded commands and shoulder trajectory, with no measurement of the end effector in between.
The error falls steadily with $k$.
The command plus a constant offset fitted on the same attempts stays near \offCoForty\,cm, and the command alone is off by \offCmd\,cm.

\begin{table}[!htbp]
  \centering
  \footnotesize
  \setlength{\tabcolsep}{3pt}
  \begin{tabular}{@{}l cccc@{}}
    \toprule
    & \multicolumn{4}{c@{}}{\makecell{Error (cm)\down\\after $k$ training attempts}} \\
    \cmidrule(l){2-5}
    Prediction & $k=5$ & $k=10$ & $k=20$ & $k=40$ \\
    \midrule
    \textbf{Learned model (ours)} & \offMbFive & \offMbTen & \offMbTwenty & \offMbForty \\
    Command plus constant offset  & \offCoFive & \offCoTen & \offCoTwenty & \offCoForty \\
    Command alone                 & \offCmd    & \offCmd   & \offCmd      & \offCmd \\
    \bottomrule
  \end{tabular}
  \caption{\textbf{The learned model predicts the end effector's path.}
  Mean distance between the predicted and the measured end-effector position over a held-out attempt's reach and hold, predicted open loop from its first measured state.
  Each cell averages over a session's attempts after the 40th, then over the nine sessions.
  Constant offset: the mean of the measured minus the commanded position over the first $k$ attempts.}
  \label{tab:suppl_model_check}
\end{table}

\paragraph{To test whether the dynamics model is needed, we compare with a model-free learner}
The model-free actor-critic of \cref{fig:rl}a also acts at every control step, yet it ends at \appMfLast\,cm against \appMbLast\,cm for our learner (\cref{sec:ablation}).
It shares our learner's correction increment, rate, bounds and acted axes, its exploration noise, and the same mode for every attempt, drawn from the same seeds.
In place of the dynamics model, it learns twin critics from the logged steps.
It is an off-policy actor-critic in the style of TD3~\cite{fujimoto2018td3}, with multi-step returns, delayed actor updates, target-policy smoothing and a behavior-cloning term toward the executed increments~\cite{fujimoto2021td3bc}.
It is updated after every attempt on all attempts so far, with targets shifted as in our goal augmentation.
Its state is the phase, the error $n_t-c_t$, the applied correction and its change over the last ten control steps.
Its reward is minus the error on the acted axes at every step of the reach and hold.

%% file: arxiv/sec/suppl/results_harness.tex
\section{Additional experiments: the harness and its agent}
\label{sec:suppl_harness_results}

\begin{figure*}[!t]
  \centering
  \includegraphics[width=\linewidth]{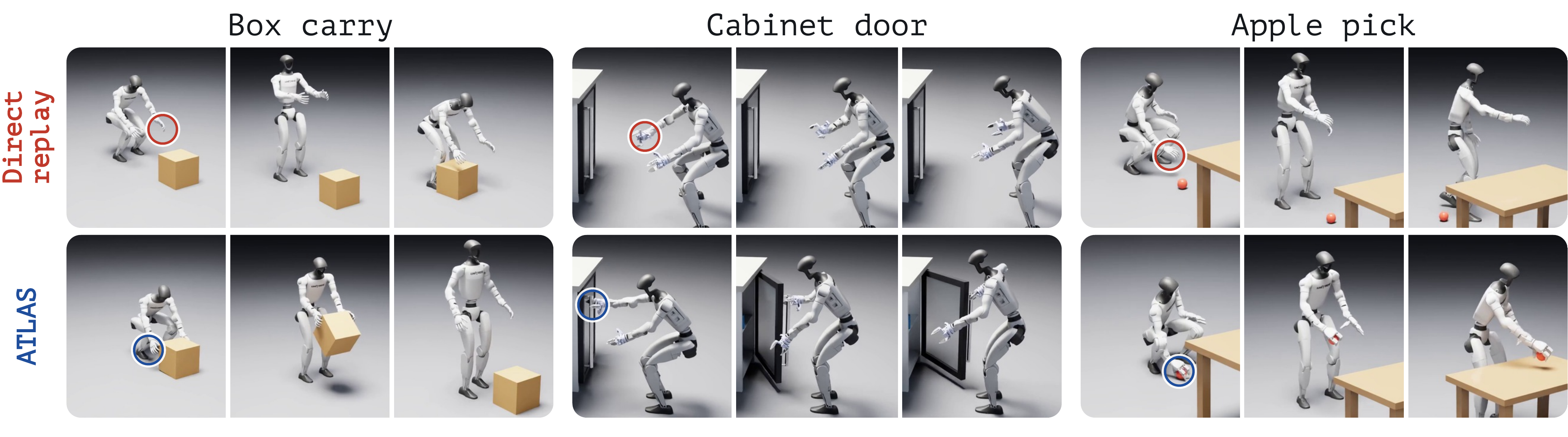}
  \caption{\textbf{The harness completes the interactions that direct replay misses.}
  Isaac Lab.
  The retargeted human motion replayed directly (top) and edited by \method (bottom), at the demonstration's contact, midway and end.
  Rings mark the end effector missing (red) or making (blue) the contact.}
  \label{fig:suppl_sim}
\end{figure*}

\paragraph{To test whether editing the motion is needed, we replay the human motion directly}
Replayed directly, the retargeted human motion fails at the contact in all three tasks of \cref{fig:suppl_sim}, and \method completes each of them.
In this baseline, the frozen tracker plays the motion as recorded in the same Isaac Lab scene as \method, with no perception and no edit of the reference.
The replayed motion closes both end effectors beside the box, stops the gripper short of the cabinet's handle and closes the end effector beside the apple.
\method aligns the reach with the perceived contact targets and runs on ScaleBFM~\cite{zeng2026scalebfm}, our default \Sys{0}.

\begin{figure*}[!t]
  \centering
  \includegraphics[width=\linewidth]{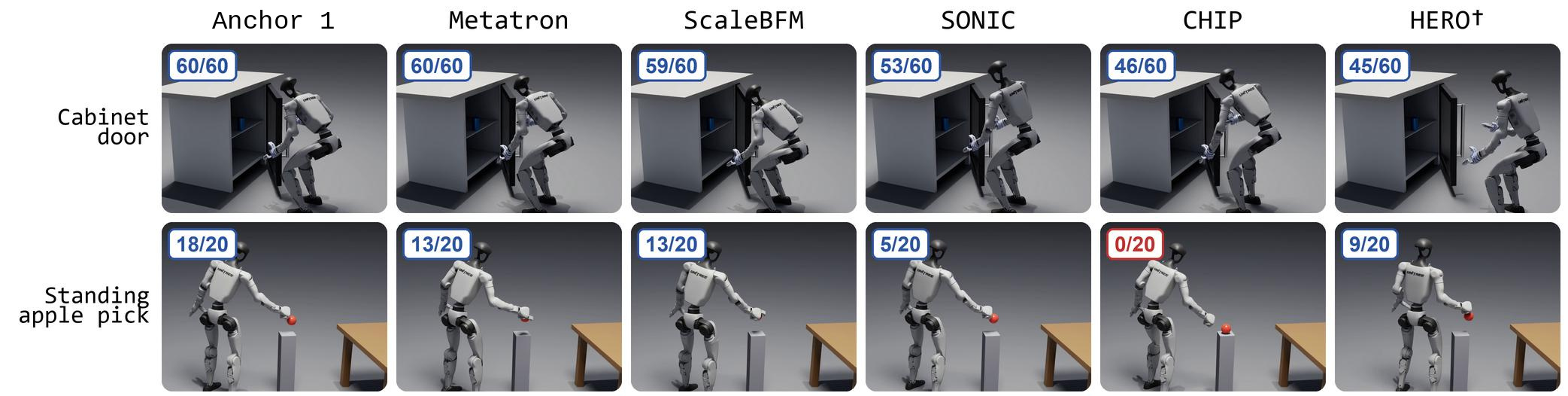}
  \caption{\textbf{The same harness on each tracker.}
  Isaac Lab, with onboard localization.
  Each cell shows one attempt at the end of the pull of the cabinet door (the door at 30\textdegree) or of the lift of the apple.
  The attempt is the first successful one after the \rlWarmup\ warm-up attempts of a separate session run to render these images, or the first after the warm-up if none succeeds.
  Badges: the tracker's success in \cref{tab:suppl_controllers}, blue when the shown attempt succeeds and red when it fails.
  HERO$^\dagger$ takes its root position from leg odometry.}
  \label{fig:suppl_controllers}
\end{figure*}

\paragraph{To test whether the harness depends on one tracker, we run it on \ctlTrackersWord\ whole-body trackers}
One fixed harness per task, with its specification and fast system, runs on each of them, and only \Sys{0} changes (\cref{tab:suppl_controllers}).
Localizing onboard, \method opens the cabinet door with every tracker and lifts the apple in a standing pick with \ctlAppleTrackersWord\ of them.
It opens the door in all \ctlDoorN\ scored attempts with Anchor~1 and with Metatron, and in \ctlBfmDoor\ with ScaleBFM.
On the door, the fast system lowers every tracker's hold error on the two axes it corrects, forward and sideways, from \ctlDoorErrWLo--\ctlDoorErrWHi\,cm before its correction to \ctlDoorErrLo--\ctlDoorErrHi\,cm after.
Besides ScaleBFM, we test SONIC~\cite{luo2025sonic} and Anchor~1~\cite{wang2026anchor}, which tracks the pelvis, wrists and ankles of the reference (\cref{fig:suppl_controllers}).
We also test three recent releases: Metatron~\cite{tessler2026metatron}, CHIP~\cite{chipcompliance2025} and HERO~\cite{dong2026hero}.
Metatron takes the torso, ankle and hand targets of the edited reference.
CHIP takes the head and palm targets and the reference's leg joints.
HERO takes palm goals through its own inverse kinematics, with its root position from leg odometry.
For SONIC, we use joint targets on the door and the three-point input (head and both hands) on the apple.
Each tracker runs three sessions on the cabinet door and one on a standing apple pick, which reuses the standing reference of \cref{fig:rl}c with the three-finger end effector; unlike the standing pick of \cref{tab:suppl_two_tasks}, the apple rests on a \ctlAppleSupportCm\,cm support and the robot localizes onboard.
A door session has \rlWarmup\ warm-up attempts and 20 scored ones; an apple session has \ctlAppleAttempts\ attempts, of which the last 20 are scored.
In the scored attempts, the actor acts without exploration.

\input{arxiv/tab/controllers}

\paragraph{To test whether the harness depends on one agent, we have \agtAgentsWord\ different agents write it}
From one human video demonstration and one prompt, an agent writes a task's harness in about a minute of model calls, without revision (\cref{tab:harness_agents}).
It annotates the interaction frames and specifies the task, and the motion-editing settings stay fixed across agents.
The \agtTasksN\ tasks span a floor pick, a box carry, a cabinet door, a candy taken from a basket, trash collected on the way to a sofa, and two stair tasks with a pick.
Claude Opus 5.5 and Haiku 5.5 write an executable harness for all \agtTasksN\ tasks, the open-weights Qwen3-VL-32B for \agtQwExec, Sonnet 5.5 for \agtSoExec\ and GPT-6.2 Astra for \agtGpExec.
In Isaac Lab with the fast system enabled, GPT-6.2 Astra's harnesses open the cabinet door in all 20 attempts after warm-up and lift the box in \agtGpBox\%.
Opus's harnesses lift the box in \agtOpBox\% and open the door in \agtOpDoor\%.
Haiku's box harness lifts the box in \agtHaBox\% of attempts and Qwen's in \agtQwBox\%.

\input{arxiv/tab/agents}

\paragraph{To test whether the agents' output is stable, we regenerate the specifications deployed on our robot}
Given only the prompt and the demonstrations, Opus reproduces the deployed specification for \agentOpusSpecK\ of \agentSpecN\ held-out tasks, Sonnet for \agentSonnetSpecK\ and Haiku for \agentHaikuSpecK.
A task counts when the \nConfigKeep\ independent generations agree and match the deployed specification in every key field (\cref{tab:suppl_spec_vocab}), with the segmenter phrases and the end state compared by meaning.
The system prompt was fixed after development on \agentDevN\ other tasks.

%% file: arxiv/tab/controllers.tex
\begin{table}[!t]
  \centering
  \footnotesize
  \setlength{\tabcolsep}{3pt}
  \renewcommand{\arraystretch}{1.0}
  \setlength{\aboverulesep}{0.3ex}
  \setlength{\belowrulesep}{0.4ex}
  \begin{tabular}{@{}l cc cc@{}}
    \toprule
    & \multicolumn{2}{c}{Cabinet door} & \multicolumn{2}{c@{}}{Standing apple} \\
    \cmidrule(lr){2-3}\cmidrule(l){4-5}
    \Sys{0} & Success\up & \makecell{Hold\\error (cm)\down} & Success\up & \makecell{Hold\\error (cm)\down} \\
    \midrule
    Anchor~1~\cite{wang2026anchor}    & \textbf{\ctlAnDoor/\ctlDoorN} & \ctlAnDoorErrW\,$\to$\,\textbf{\ctlAnDoorErr} & \textbf{\ctlAnApple/\ctlAppleN} & \ctlAnAppleErrW\,$\to$\,\ctlAnAppleErr \\
    Metatron~\cite{tessler2026metatron}  & \textbf{\ctlMeDoor/\ctlDoorN} & \ctlMeDoorErrW\,$\to$\,\ctlMeDoorErr & \ctlMeApple/\ctlAppleN & \ctlMeAppleErrW\,$\to$\,\textbf{\ctlMeAppleErr} \\
    ScaleBFM~\cite{zeng2026scalebfm} & \ctlBfmDoor/\ctlDoorN & \ctlBfmDoorErrW\,$\to$\,\textbf{\ctlBfmDoorErr} & \ctlBfmApple/\ctlAppleN & \ctlBfmAppleErrW\,$\to$\,\ctlBfmAppleErr \\
    SONIC~\cite{luo2025sonic}        & \ctlSonDoor/\ctlDoorN & \ctlSonDoorErrW\,$\to$\,\ctlSonDoorErr & \ctlSonApple/\ctlAppleN & \ctlSonAppleErrW\,$\to$\,\ctlSonAppleErr \\
    CHIP~\cite{chipcompliance2025}          & \ctlChDoor/\ctlDoorN & \ctlChDoorErrW\,$\to$\,\ctlChDoorErr & \ctlChApple/\ctlAppleN & \ctlChAppleErrW\,$\to$\,\ctlChAppleErr \\
    HERO~\cite{dong2026hero}$^\dagger$ & \ctlHeDoor/\ctlDoorN & \ctlHeDoorErrW\,$\to$\,\ctlHeDoorErr & \ctlHeApple/\ctlAppleN & \ctlHeAppleErrW\,$\to$\,\ctlHeAppleErr \\
    \bottomrule
  \end{tabular}
  \caption{\textbf{The same harness runs on \ctlTrackersWord\ whole-body trackers.}
  Isaac Lab, with onboard localization; on each task, every tracker runs the same harness, with its specification and fast system.
  Success: the door opened at least 30\textdegree, or the apple lifted and held, over the scored attempts.
  Hold error, on the axes the correction acts on (forward and sideways at the door; forward, sideways and down at the apple): median over a session's warm-up attempts, before the fast system corrects, $\to$ median over its scored attempts; both averaged over sessions.
  $^\dagger$With its root position from leg odometry and the simulator's joint control in place of its own.
  Apple error: attempts with the apple on its support at contact.}
  \label{tab:suppl_controllers}
\end{table}

%% file: arxiv/tab/agents.tex
\begin{table}[t]
  \centering
  \footnotesize
  \setlength{\tabcolsep}{3.5pt}
  \renewcommand{\arraystretch}{1.0}
  \setlength{\aboverulesep}{0.3ex}
  \setlength{\belowrulesep}{0.4ex}
  \begin{tabular}{@{}l cccc@{}}
    \toprule
    Agent & \makecell{Executable\\(of \agtTasksN)\up} & \makecell{Time\\(s)\down} & \makecell{Box lift\\(\%)\up} & \makecell{Cabinet door\\(\%)\up} \\
    \midrule
    Claude Haiku 5.5  & \textbf{\agtHaExec} & \agtHaTime & \textbf{\agtHaBox} & \agtHaDoor \\
    Claude Sonnet 5.5 & \agtSoExec & \agtSoTime & \agtSoBox & no harness \\
    Claude Opus 5.5   & \textbf{\agtOpExec} & \agtOpTime & \agtOpBox & \agtOpDoor \\
    GPT-6.2 Astra     & \agtGpExec & \textbf{\agtGpTime} & \agtGpBox & \textbf{\agtGpDoor} \\
    Qwen3-VL-32B      & \agtQwExec & \agtQwTime & \agtQwBox & no harness \\
    \bottomrule
  \end{tabular}
  \caption{\textbf{Different agents write executable harnesses from one human video demonstration and one prompt.}
  Each agent writes once, without revision; these single writes are separate from the fixed harnesses run in \cref{tab:suppl_controllers}, and the motion-editing settings are fixed.
  Executable: passes the parsing, vocabulary and frame checks (within the recording and mutually consistent), with no refusal.
  Time: mean wall time of the agent's calls per harness.
  Box lift and cabinet door: success in Isaac Lab with our learner, over 20 attempts after \rlWarmup\ warm-up attempts; the box lifted and held, or the door opened at least 30\textdegree.
  No harness: the agent's door answer is not executable; Sonnet's cut of the video fails the frame check, and Qwen3-VL declines, judging the demonstration unfit for a door.
  Haiku's door harness pulls with the left hand, whereas the person pulls with the right, and opens the door in no attempt.}
  \label{tab:harness_agents}
\end{table}

%% file: arxiv/sec/suppl/results_slow.tex
\section{Additional experiments: the slow system}
\label{sec:suppl_slow_results}
The slow system fixes two failures that a command correction cannot: the segmenter misses the object because the phrase that names it does not fit the scene, or the robot judges an attempt a success although its end effector is empty.
In both, the agent reads the robot's saved head-camera frames and rewrites text: the object phrase, or the check that judges success (\cref{sec:slow}).
The first two experiments test each rewrite offline, replayed on later frames of real-robot runs, and the last two run the slow system online in simulation.
The first three give the full results behind \cref{sec:exp_slow}.

\paragraph{To test whether the rewritten phrase fixes missed detections, we replay it on later real-robot frames}
With the phrase in force, the segmenter often missed the object in \slowRunsWord\ real-robot runs.
The agent proposes new phrases from each run's early frames and keeps the one that scores best on its own labels of those frames.
The segmenter is then replayed with that phrase on each run's later frames, \slowRecallN\ in all with the object in view.
Recall, the share of these frames on which the segmenter finds the object, rises from \slowRecallOffD\% to \slowRecallOnD\%, with wrong-object readings at \slowWrongOnD\% (\slowWrongOffD\% with the phrase in force; \cref{tab:slow}a).
A segmenter threshold tuned on the same early frames reaches \slowRecallThrD\%, below the agent's phrase (Holm $p=\slowRecallThrHolmP$), and a paraphrase written without the frames reaches \slowRecallParaD\% but boxes the wrong object on \slowWrongParaD\% of the frames.
With Sonnet 5.5 or Haiku 5.5 as the agent, recall is \agentSonnetRecallD\% and \agentHaikuRecallD\% (\cref{tab:slow}b); Haiku keeps the phrase in force on \slowHaikuKeptRunsWord\ of the \slowRunsWord\ runs.
The segmenter finds the object when its box contains the object's point as labeled in \cref{sec:suppl_protocol}; a wrong-object reading is a box that misses that point, exceeds twice the object's extent, or falls on a frame without the object.

\input{arxiv/tab/slow}

\paragraph{To test whether the rewritten success check fixes wrong judgments, we replay it on later real-robot attempts}
The robot judges every attempt itself: a false success ends a trial that failed, and a false miss costs a retry.
The agent rewrites the check from the earlier attempts of each real-robot run.
Each check in \cref{tab:suppl_slow_indicator} is then replayed on \slowIndN\ later attempts and compared with blind labels, read afterwards from each attempt's frames and from the next approach's first look at the object's rest spot.
The agent's check cuts false successes by two thirds, from \slowIndFalseSuccOffK\ to \slowIndFalseSuccOnK\ of the \slowIndNotHeldN\ attempts that held nothing, and disagreements with the labels from \slowIndDisagreeOff\ to \slowIndDisagreeOn\ (Holm $p=\slowIndDisagreeHolmP$; \cref{tab:suppl_slow_indicator}); false misses rise from \slowIndFalseMissOffK\ to \slowIndFalseMissOnK\ of the \slowIndHeldN\ attempts that held the object.
A logistic regression fitted to the labels of the earlier attempts, and the agent without frames, suppress more false successes but reject most of the held objects.

\begin{table}[!t]
  \centering
  \footnotesize
  \setlength{\tabcolsep}{3.2pt}
  \begin{tabular}{@{}l ccc@{}}
    \toprule
    Success check & \makecell{False\\success (\%)\down} & \makecell{False\\miss (\%)\down} & \makecell{Disagree-\\ments\down} \\
    \midrule
    Rule in force                 & \slowIndFalseSuccOff\ (\slowIndFalseSuccOffK/\slowIndNotHeldN)     & \slowIndFalseMissOff\ (\slowIndFalseMissOffK/\slowIndHeldN)     & \slowIndDisagreeOff \\
    Logistic regression           & \slowIndFalseSuccLogit\ (\slowIndFalseSuccLogitK/\slowIndNotHeldN) & \slowIndFalseMissLogit\ (\slowIndFalseMissLogitK/\slowIndHeldN) & \slowIndDisagreeLogit \\
    Agent without frames          & \slowIndFalseSuccRows\ (\slowIndFalseSuccRowsK/\slowIndNotHeldN)   & \slowIndFalseMissRows\ (\slowIndFalseMissRowsK/\slowIndHeldN)   & \slowIndDisagreeRows \\
    \makecell[l]{\textbf{Agent with frames}\\\textbf{(ours)}} & \slowIndFalseSuccOn\ (\slowIndFalseSuccOnK/\slowIndNotHeldN) & \slowIndFalseMissOn\ (\slowIndFalseMissOnK/\slowIndHeldN) & \textbf{\slowIndDisagreeOn} \\
    \bottomrule
  \end{tabular}
  \caption{\textbf{The agent's revised success check cuts false successes by two thirds.}
  Checks replayed on \slowIndN\ held-out real-robot attempts and scored against blind hindsight labels; the logistic regression is fitted to the labels of the earlier attempts.
  False success: a success judged among the \slowIndNotHeldN\ attempts that held nothing; false miss: a miss judged among the \slowIndHeldN\ that held the object; disagreements: their sum, of \slowIndN.}
  \label{tab:suppl_slow_indicator}
\end{table}

\begin{figure}[!t]
  \centering
  \includegraphics[width=\linewidth]{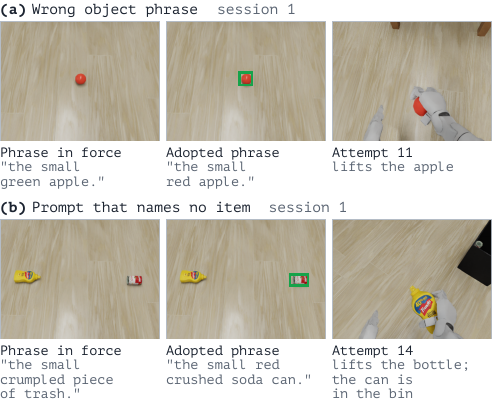}
  \caption{\textbf{Under the phrase the review writes, the segmenter finds the object and the robot picks it up.}
  Head-camera frames from session~1 of the two simulated runs.
  In each row, the first two pictures are one frame saved in attempt~2 and replayed by the review under the phrase in force and under the phrase it adopts; the box is the segmenter's.
  (a)~The wrong object phrase (\cref{tab:suppl_slow_closed}).
  (b)~The prompt that names no item (\cref{tab:suppl_slow_trash}).}
  \label{fig:suppl_slow_sim}
\end{figure}

\paragraph{To test whether the slow system repairs a wrong object phrase online, we start a simulated pick with one}
The phrase is ``the small green apple.'', while the scene's apple is red, so the segmenter does not find it (\cref{fig:suppl_slow_sim}a).
Without the slow system, nothing revises the phrase, and the robot lifts the apple in \slowLoopOffK\ of \slowLoopN\ attempts.
With it, every session adopts ``the small red apple.'' within two review cycles, and the robot lifts the apple in \slowLoopOnK\ of \slowLoopN\ attempts, \slowLoopAfterK\ of the \slowLoopAfterN\ after the switch; the retry in place recovers each of the \slowLoopAfterMissesWord\ misses (\cref{tab:suppl_slow_closed}).
Each condition runs \slowLoopSeedsWord\ sessions of \slowLoopAttemptsPerSeed\ attempts with the fast system enabled, and the simulator's state scores every attempt.
A review cycle takes \slowLoopCycleSLo\ to \slowLoopCycleSHi\,s while the robot keeps acting on the phrase in force, as on the real robot.
The agent sees the prompt, the phrase in force, the segmenter's counts and saved head-camera frames, but never the simulator's state; a candidate phrase is adopted under the acceptance rule of \cref{sec:suppl_agent}, at a Holm-corrected $p<0.05$.

\begin{table}[!t]
  \centering
  \footnotesize
  \setlength{\tabcolsep}{4.5pt}
  \begin{tabular}{@{}c ccccc@{}}
    \toprule
    Session & \makecell{Off,\\lifted} & \makecell{On,\\lifted} & Cycles & Switch & \makecell{After the\\switch, lifted} \\
    \midrule
    1   & \slowLoopSeedOneOffK/\slowLoopAttemptsPerSeed   & \slowLoopSeedOneOnK/\slowLoopAttemptsPerSeed   & \slowLoopSeedOneCycles   & \slowLoopSeedOneSwitch   & \slowLoopSeedOneAfterK/\slowLoopSeedOneAfterN \\
    2   & \slowLoopSeedTwoOffK/\slowLoopAttemptsPerSeed   & \slowLoopSeedTwoOnK/\slowLoopAttemptsPerSeed   & \slowLoopSeedTwoCycles   & \slowLoopSeedTwoSwitch   & \slowLoopSeedTwoAfterK/\slowLoopSeedTwoAfterN \\
    3   & \slowLoopSeedThreeOffK/\slowLoopAttemptsPerSeed & \slowLoopSeedThreeOnK/\slowLoopAttemptsPerSeed & \slowLoopSeedThreeCycles & \slowLoopSeedThreeSwitch & \slowLoopSeedThreeAfterK/\slowLoopSeedThreeAfterN \\
    \midrule
    All & \slowLoopOffK/\slowLoopN & \slowLoopOnK/\slowLoopN & \slowLoopCyclesAll & -- & \slowLoopAfterK/\slowLoopAfterN \\
    \bottomrule
  \end{tabular}
  \caption{\textbf{The slow system rescues a pick that never succeeds without it.}
  Lifted: attempts that lift the apple, by the simulator's state, of \slowLoopAttemptsPerSeed\ per session and condition.
  Cycles: review cycles run with the slow system on; the last one keeps ``the small red apple.''
  Switch: the first attempt on the new phrase.}
  \label{tab:suppl_slow_closed}
\end{table}

\paragraph{To test the slow system on a prompt that names none of the items, we have the robot clear a simulated floor of trash}
With the slow system, the robot puts every item in the bin in every session, and nothing is picked up without it (\cref{tab:suppl_slow_trash}).
Under the prompt ``Pick up the trash and put it in the bin.'', the floor holds a red and white can, a yellow plastic mustard bottle and a bin.
The human video demonstration is a banana pick and supplies only the motion.
The prompt names no item, and the agent's first phrase, ``the small crumpled piece of trash.'', matches none of them (\cref{fig:suppl_slow_sim}b).
With the slow system on, the first review names the can, which goes into the bin at attempt \trashFirstAt, the first under the new phrase.
The second review names the bottle, and the floor is clear after \trashFloorAtLo\ or \trashFloorAtHi\ attempts.
With it off, the robot keeps that phrase, and all \trashAttempts\ attempts of every session end before the contact.
On the real robot, the slow system replaces the phrase ``the trash.'' with one phrase per object (\cref{fig:slow_real}).

\begin{table}[!t]
  \centering
  \footnotesize
  \setlength{\tabcolsep}{5pt}
  \begin{tabular}{@{}c cccc@{}}
    \toprule
    Session & \makecell{Off,\\cleared} & \makecell{On,\\cleared} & \makecell{Cleared\\by attempt} & \makecell{Cycles\\run/kept} \\
    \midrule
    1   & \trashSeedOneOffK/\trashItems   & \trashSeedOneOnK/\trashItems   & \trashSeedOneFirstAt, \trashSeedOneFloorAt     & \trashSeedOneCycles/\trashSeedOneKept \\
    2   & \trashSeedTwoOffK/\trashItems   & \trashSeedTwoOnK/\trashItems   & \trashSeedTwoFirstAt, \trashSeedTwoFloorAt     & \trashSeedTwoCycles/\trashSeedTwoKept \\
    3   & \trashSeedThreeOffK/\trashItems & \trashSeedThreeOnK/\trashItems & \trashSeedThreeFirstAt, \trashSeedThreeFloorAt & \trashSeedThreeCycles/\trashSeedThreeKept \\
    \midrule
    All & \trashOffK/\trashN & \trashOnK/\trashN & -- & \trashCyclesAll/\trashCyclesKeptAll \\
    \bottomrule
  \end{tabular}
  \caption{\textbf{The slow system clears a floor of trash.}
  Cleared: items in the bin by the simulator's state, of \trashItems\ per session and condition.
  Cleared by attempt: with the slow system on, the attempts after which the first item is in the bin and the floor is clear.
  Cycles: review cycles run and kept with the slow system on.
  The kept phrases, in the order they take force: session~1, ``the small red crushed soda can.'' and ``the yellow plastic mustard bottle.''; session~2, ``the small red and white crushed can.'' and ``the small yellow plastic bottle.''; session~3, ``the small red and white tin can.'' and ``the small yellow plastic mustard bottle.''}
  \label{tab:suppl_slow_trash}
\end{table}

%% file: arxiv/tab/slow.tex
\begin{table}[t]
  \centering
  \footnotesize
  \newcommand{\nv}[1]{\tablenum[table-format=2.1,reset-text-shape=false]{#1}}
  \newcommand{\nvu}[1]{\settowidth{\dimen0}{00.0}\makebox[\dimen0][r]{\underline{#1}}}
  \begin{tabular}{@{}l cc@{}}
    \toprule
    & \makecell{Recall\\(\%)\up} & \makecell{Wrong\\object (\%)\down} \\
    \midrule
    \multicolumn{3}{@{}l}{\textit{(a) Who revises the phrase}} \\
    Phrase in force (control)         & \nv{\slowRecallOffD}       & \nv{\slowWrongOffD} \\
    Threshold tuned                   & \nv{\slowRecallThrD}       & \nvu{\slowWrongThrD} \\
    Paraphrase without frames         & \nvu{\slowRecallParaD}     & \nv{\slowWrongParaD} \\
    \textbf{Agent with frames (ours)} & \textbf{\nv{\slowRecallOnD}} & \textbf{\nv{\slowWrongOnD}} \\
    \midrule
    \multicolumn{3}{@{}l}{\textit{(b) Which model is the agent}} \\
    Claude Haiku 5.5                  & \nv{\agentHaikuRecallD}    & \textbf{\nv{\agentHaikuWrongD}} \\
    Claude Sonnet 5.5                 & \nvu{\agentSonnetRecallD}  & \nv{\agentSonnetWrongD} \\
    \textbf{Claude Opus 5.5}          & \textbf{\nv{\agentOpusRecallD}} & \nvu{\agentOpusWrongD} \\
    \bottomrule
  \end{tabular}
  \caption{\textbf{The agent quadruples recall with almost no wrong readings.}
  Revisions come from early frames of \slowRunsWord\ real-robot runs where the phrase failed and are scored on later ones (\slowRecallN\ with the object in view).
  Bold best, underline second among the revising methods.}
  \label{tab:slow}
\end{table}

%% file: arxiv/sec/suppl/spec.tex
\section{Method details: the task specification and motion editing}
\label{sec:suppl_spec}

\paragraph{From the video to the reference}
The person's motion, recovered from the demonstration's video by GVHMR~\cite{gvhmr2024} and retargeted to the G1 by GMR~\cite{gmr2025}, is cut by one model call.
It reads video frames, one per \cutFrameStepS\,s, beside a table measured on the retargeted motion, one row per \cutTableStepS\,s, and names the demonstrator's working hands, the grasp and release frames and the segments: a \emph{pick} and, for a release in a later stance, a \emph{place}, each by its \emph{start}, \emph{contact}, \emph{settle} and \emph{end}.
The \emph{held point}, the working end effector's contact point from grasp to release, is the path the affordance moves onto the object.

\paragraph{What the agent reads and answers}
The agent answers once per task, before the first attempt, under a fixed system prompt and without tools.
It sees the task's prompt, the robot's end effectors in words (fingers, a jaw or a rigid palm, and how each closes) and the demonstrations it may choose from as measured facts under neutral identifiers that name no object or task: the working hands, the action axes, the grasp and release stances, the held point's rise and the contact height.
It answers in the closed vocabulary of \cref{tab:suppl_spec_vocab} with a one- or two-sentence reason, or declines.

\begin{table*}[!t]
  \centering
  \tabstyle
  \begin{tabular}{@{}l >{\raggedright\arraybackslash}p{0.80\linewidth}@{}}
    \toprule
    Field & Values, and what the robot does with each \\
    \midrule
    demonstration & the identifier of one demonstration, or a decline \\
    object, destination & a segmenter phrase each, \emph{the $\langle$color$\rangle$ $\langle$material or size$\rangle$ $\langle$common name$\rangle$.}, never where it sits; the object phrase names the part the end effector holds; the destination may be none, and with one the robot carries the object there and lets go at the demonstrated standoff \\
    contact.feature & \texttt{face\_centre}, \texttt{edge}: the end effector arrives on the upper face; \texttt{lateral\_surface}: it closes around the object, the site is its measured center; \texttt{handle\_inner\_surface}, \texttt{bar}, \texttt{lip}: the center of the part the segmenter finds, two end effectors along a bar's length as far apart as the demonstrator's wrists \\
    contact.relation & \texttt{opposite}: two end effectors on either side along the closing line; \texttt{same}: one, at its feature's site; \texttt{none}: every one at the measured center \\
    closure.direction & \texttt{toward\_each\_other}: two opposite end effectors close until the object stops them; \texttt{away\_from\_each\_other}: they move apart, nothing closing on the object; \texttt{along\_axis\_into\_object}: a push or press; \texttt{along\_axis\_out\_of\_object}: a pull; \texttt{enclose}: fingers or jaw close around the object, the wrist still; \texttt{none} \\
    closure.axis & one of the demonstration's action axes: \texttt{fore}, \texttt{grasp} (across the object) or \texttt{up} \\
    closure.grip\_force\_n & newtons each end effector presses when the closure carries the object's weight, or null for a push, pull or press; it selects the progress indicator's geometric checks \\
    outcome.end\_state & one present-tense sentence of what the head camera sees once the motion has ended, if the request has succeeded \\
    measured, not asked & the object's size, shape and axes and the closing line, from the mask and the depth when the robot looks \\
    \bottomrule
  \end{tabular}
  \caption{\textbf{The closed vocabulary of the task specification.} Each value maps to one fixed behavior of the robot. The segmenter phrases, the site sentences (one per end effector) and the reason are free text and the grip force a free number; none of them enters the agreement signature beyond whether a force is given.}
  \label{tab:suppl_spec_vocab}
\end{table*}

\paragraph{The agreement rule}
A multi-demonstration request is first split into ordered steps, each a demonstration and a destination kind.
The split and every step's setting are asked \nConfigKeep\ times in isolated calls and compared by their \emph{signature}: the split's ordered steps; a setting's demonstration, destination, contact relation and feature class, closure direction and axis, whether a force is given, and end state.
The specification is written only when all \nConfigKeep\ signatures agree, or a strict majority where the operator accepts one; otherwise a disagreement writes nothing.
At test time only the slow system revises it (\cref{sec:slow}).

\paragraph{The segmenter phrases}
SAM~3~\cite{sam3_2025} has not seen the scene, so each phrase gives the name and the look together, since either alone misses objects the other finds.

\paragraph{Placing the reference and aligning the wrist}
At the handoff, the reference is turned in yaw and shifted on the floor so that its pelvis at the interaction start coincides with the robot's pelvis as the tracker reports it (\cref{sec:edit}).
Only the interacting wrist's target receives the offset $p^\star-c^{\mathrm{ref}}_\tau$ of \cref{eq:contact_alignment}, from the demonstrated contact point to the perceived site; the tracker generates the whole-body motion from the edited reference.
The wrist keeps the demonstrated orientation unless the object's geometry or the end effector calls for a turn.

\paragraph{Updates during the reach}
Without external localization, the wrist target follows the object and the body during the reach (\cref{sec:edit}).
Consistent head-camera readings re-anchor the site during the crouch, and the support leg's kinematics compensate the pelvis drift; both enter scaled by $\alpha_t$.
At the closure decision these updates stop and the shift reached so far is kept.
The learned correction $u_t$ is added on top, in the attempt frame (\cref{sec:fast}).

\paragraph{The progress indicator}
With a grip force the indicator is geometric: the closure at the lift's end and whether the head camera sees the object above its rest height (\cref{sec:suppl_agent}); without one, a closure on nothing ends the attempt at once.
From the contact on, one VLM judge for every task, \judgeModel, reads the head-camera frame with the task's words, and two consecutive looks naming a failure (the object lost or, in a carry or set-down, still where it rested) end the attempt.
After the last frame it records whether the end-state sentence holds; success is scored as in \cref{sec:suppl_sim,sec:suppl_protocol}.

%% file: arxiv/sec/suppl/learner.tex
\section{Method details: the fast system}
\label{sec:suppl_learner}

\paragraph{Window}
In each attempt the fast system acts from reach onset (\cref{sec:edit}) through the hold at the contact to the step $T$ that decides the close; the correction then ramps to zero (\cref{tab:suppl_learner_settings}).
The target $n_t$ never contains the correction, so the learner cannot lower its error by moving its own target.
The tracker's input describes each end effector by its wrist link, rigidly offset from the contact point where $n_t$ and $c_t$ are taken, and the correction shifts only that link's position in the input.

\paragraph{Plant and dynamics model}
The plant maps the command, the harness's plus the applied correction, to $c_t$, a fixed contact point of the end effector, read by forward kinematics on the robot and from the simulated body's pose in simulation.
The dynamics model splits it as $c_t=b_t+h_t$, takes the shoulder $b_t$ as exogenous and replays it as recorded, and predicts $h_t$, the end effector relative to the shoulder.
Each member of \cref{eq:model} sums a linear map, a SiLU network and a fixed random prior of the same structure~\cite{osband2018rpf}, and predicts the change of $h$ from the phase, the recent and look-ahead commands relative to $b_t$, and $h_t$ with its last change.
Each update fits the ensemble by a free run over every attempt's window from its first measured $h$, SuperMimic's multi-step loss~\cite{liao2026supermimic}, warm-started from the session's previous fit.
On the real robot (\cref{tab:real_ablation}), the fit also carries, per acted axis, a data-free prior $\mathcal N(1,1)$ on the gain by which the end effector's hold position moves per unit of correction, so that the sign of the response is not decided by the few warm-up attempts alone; the session's own hold data update it.

\paragraph{Reward and actor}
The reward of \cref{eq:reward} takes the error only on the axes the correction acts on; its cost is the weighted squared change of the correction, in units of the rate.
Per end effector, the actor $\pi_\phi$ of \cref{eq:actor} reads the model's input, the error $n_t-c_t$ now and ahead, the applied correction and the correction still in flight, and outputs an increment squashed radially into the ball the tracker's correction channel admits per step; $u_t$ sums the increments, clipped to the bound and to a length cap.
It maximizes the undiscounted reward summed over the ensemble and over every past attempt that began at zero correction, reached the hold and never met the floor reflex, by which the correction channel holds back a downward correction once the end effector touches the floor.
Each such attempt is replayed under the current actor, with its target shifted by a goal displacement from the session (\cref{sec:fast}) and its end-effector reading noised at the model's held-out one-step error.
The gradient runs through the fixed model and through twins of the robot's squash, bound, cap, measured delay and correction channel, one-sided where bound or cap saturate, so a saturated actor is pulled back without wind-up.

\paragraph{Warm-up, modes and updates}
The first \rlWarmup\ attempts of a session are warm-up: a zero correction plus Gaussian exploration increments, the actor not read, so the dynamics model sees the correction act before the actor does.
After them, in the sessions of \cref{fig:rl}a and of \cref{fig:suppl_ablation,tab:suppl_two_tasks}, a coin from the learner's own generator deals each attempt's mode: \emph{zero} (no correction, the control) with probability \rlArmZeroP, \emph{mean} (the actor without exploration) with \rlArmMeanP, and \emph{explore} (the actor exploring) otherwise.
In \cref{tab:real_ablation} and at the random placements (\cref{sec:suppl_sim}), no coin is dealt: every attempt after the warm-up runs the actor without exploration, and the control runs in sessions of its own.
Every attempt but a fall, first or retry and whatever its mode, trains the learner.
The robot sends its logged steps to the PC, which updates both networks in the background on all the attempts so far; the robot never waits for it.
An update is accepted if its parameters are finite and the twin of the correction channel left every step of the model's rollouts unchanged.
Every attempt starts with the newest accepted actor, under the same hard bound, rate and cap as in the model's rollouts.
No success label or held-out score selects an update.
The correction acts forward and sideways and, on the floor picks of \cref{tab:real_ablation} and \cref{fig:rl}a, at the random placements (\cref{sec:suppl_sim}) and in the sessions of \cref{fig:suppl_ablation,tab:suppl_two_tasks}, downward as well (\cref{tab:suppl_learner_settings}).

\begin{table*}[!t]
  \centering
  \footnotesize
  \begin{tabular}{@{}lll@{}}
    \toprule
    Part & Setting & Value \\
    \midrule
    Window & hold at the contact; ramp to zero after the close & $\ge$\rlHoldS\,s; $\ge$\rlRampFrames\ control steps \\
    Dynamics model & members; hidden layers (SiLU); inputs per end effector & \rlMembers; $2\times\rlHidden$; \rlModelInDim \\
     & look-ahead frames read; previous commands read & \rlLookSlots; \rlHistPrev \\
     & Adam steps per update; learning rate & \rlModelSteps; \rlLrModel \\
    Reward & error unit; weight of the correction's change & \rlErrUnitCm\,cm; \rlSmoothW \\
    Actor & hidden layers; inputs per end effector & $2\times\rlHidden$; \rlStateDim\ (67 with the downward axis) \\
     & error read ahead & \rlNeedLo\ and \rlNeedHi\ control steps \\
     & Adam steps per update; learning rate & \rlActorSteps; \rlLrActor \\
    Limits & bound forward; sideways; downward; rate & \rlBoundXHiCm\,cm; $\pm$\rlBoundYCm\,cm; \rlBoundZCm\,cm; \rlRateMm\,mm per step \\
    Exploration & increment s.d.\ per step and axis & \rlExploreMm\,mm (\rlExploreFrac\ of the rate) \\
    Modes & warm-up attempts; zero / mean probability & \rlWarmup; \rlArmZeroP\ / \rlArmMeanP \\
    \bottomrule
  \end{tabular}
  \caption{\textbf{The fast system's settings,} on the robot and in simulation. In the sessions dealt by a coin, the rest of its probability goes to exploration.}
  \label{tab:suppl_learner_settings}
\end{table*}

%% file: arxiv/sec/suppl/agent.tex
\section{Method details: the slow system}
\label{sec:suppl_agent}

\paragraph{What the agent reads}
The slow system is the VLM agent of \cref{sec:spec}, running on the PC beside the robot (\cref{sec:slow}).
For each attempt it receives one row the robot measures itself: the progress indicator's outcome, where the walk ended, the object's measured extent, the end effector's gap to the object at the close, the contact miss $c_T-n_T$, and the correction with its bound hits.
It also sees success tabulated by first attempt and retry, by whether the correction acted, and by contact miss, gap and perceived height; the \slowRowsShown\ newest successes and failures each; and the close and lift frames of the \slowFrames\ newest of each, never the simulator's state.

\paragraph{When it acts}
The agent proposes at most one revision at a time, never while a scored one is pending; the robot keeps acting on the specification in force while the agent works and loads the newest phrase before each attempt, so an accepted phrase takes effect at the next attempt (\cref{sec:slow}).

\paragraph{What it may change: the segmenter phrases}
The object phrase, the phrase of the part the end effector holds (a handle bar, for example) and the destination phrase stay in the specification's template, ``the \textless color\textgreater\ \textless material or size\textgreater\ \textless common name\textgreater.'', with no location words.
The agent is shown the frames where the object was in view but unfound or its mask fell elsewhere, writes two or three alternatives, each with the reason it reads from the frames, and the code scores them.
\emph{Recall} is the share of frames with the object in view whose mask's bounding box contains the object's point, taken from the end effector's contact point by forward kinematics at a success or from a point a VLM labeler marks on the frame, with an extent at most twice the object's.
\emph{Wrong object} counts masks on the robot's own links, projected into the head camera by forward kinematics, on the operator, or with an extent above twice the object's.
The incumbent and each candidate are replayed on the same saved frames, with fresh segmenter sessions; in the simulated closed loop (\cref{tab:suppl_slow_closed}), a candidate is kept only if its recall on the frames it was not proposed from rises, by an exact McNemar test on the discordant frames, Holm-corrected over the candidates, and its wrong-object rate does not.
On the real robot (\cref{fig:slow_real}), the agent writes one phrase for each object of the task it sees, each checked on the saved frames, and the harness grounds them in turn; a revised phrase is kept if its detections of the object, net of wrong-object detections, exceed those of the phrase in force.

\paragraph{What it may change: the progress indicator}
The indicator's rule says which evidence counts, closure at the lift's end, head-camera readings above the rest height, a reading inside the end effector, the palm gap in a carry, and how they combine; its VLM query is the end-state question, asked only where the head camera can see the answer.
Each attempt gets a hindsight label from its close, lift and carry frames and the next approach's first look at the rest spot: a \emph{false success} is an empty end effector in the carry or the object still at its rest spot at the next look, a \emph{false miss} the object in the end effector after a judged miss.
A revision is replayed deterministically on every logged attempt, the judge re-asked on the saved frames, and kept only if, on the attempts it was not proposed from, false successes do not rise and disagreements fall; a false success loses a whole trial, a false miss costs one retry.

%% file: arxiv/sec/suppl/sim.tex
\section{Simulation protocol}
\label{sec:suppl_sim}

\paragraph{Isaac Lab in place of the robot}
Simulated runs execute the robot's harness and learner code on the simulator's own runner, with nothing pretrained for the task.
The simulator replaces the body and its sensors: a G1 with the task's end effector (for the Dex3 fingers, the contact point is the centroid of the closed fingertips), the rendered head camera and depth image, the joint state and the root pose.
A pick succeeds if the object ends the lift at least \simLiftCm\,cm above where it lay and within \simHandObjCm\,cm of the end effector's contact point.
This ground-truth score, not the robot's progress indicator, decides the retry.

\paragraph{Floor picks}
The simulated floor picks take an apple (\cref{fig:rl}a and \cref{fig:suppl_ablation}) or, at the random placements (below), a banana, a can or an apple off flat ground with the Dex3 fingers on the G1's right wrist.
All three use one floor-pick demonstration, in which a person walks up, squats, takes a banana off the floor with the right hand, carries it and sets it down; the can and the apple each get their own segmenter phrase and affordance.

\paragraph{Cabinet door, box carry and standing apple pick (\cref{tab:suppl_controllers,tab:harness_agents})}
The cabinet, the box and the apple's \ctlAppleSupportCm\,cm support stand at fixed places in their scenes, and the robot walks from its start to its station.
It finds the handle, the box or the apple with the head camera and the segmenter.
At the door and the apple it localizes by its own odometry, and at the box from the simulator's root pose.
In \cref{tab:suppl_controllers}, the robot reads no pose from the simulator.
Every door and apple attempt starts from the reset scene, and a missed box lift is retried in place.
The door succeeds when its hinge opens at least 30\textdegree\ within the demonstrated pull, and the box when it is lifted and held between the palms; the apple follows the pick rule above.
On the apple, the correction also acts downward, by up to \ctlAppleDownCm\,cm.
While the open hand holds at the contact, the simulator keeps the apple where it is, and when the fingers close it attaches the apple to the end effector if the contact point is within one apple diameter of the apple's center.

\paragraph{The learners}
All act through the fast system's window, bounds and rate (\cref{sec:fast}).
Let $e_i$ be the mean end-effector error $c_t-n_t$ over the hold of attempt $i$ and $u_i$ the correction applied in it.
The \emph{constant offset} sends $u=-\overline{(e-u)}$ for the rest of the session, the bar being the mean over the warm-up attempts.
\emph{Recursive least squares} fits $e_i=\beta+G\,u_i$ per axis, a bias and a gain, after every attempt and sends $u=-\hat\beta/\max(\hat G,0.3)$ on the next, clipped to the correction's bounds.
Its priors, noise and forgetting ($1-1/80$) are set from the robot's measured execution statistics.
\emph{Fixed-gain feedback}, run on the robot only, sends $u_{i+1}=u_i-e_i$, the last correction minus the last error, as if the gain were one.
All share the fast system's warm-up and hold their correction constant within an attempt.
The \emph{model-based} learner is the fast system, unchanged.
The \emph{model-free actor-critic}, in \cref{fig:rl}a only, shares the fast system's correction increment and per-attempt modes and learns twin critics in place of the dynamics model (\cref{sec:suppl_fast_results}).

\paragraph{One placement, perceived (\cref{fig:rl}a)}
Here the apple lies 0.6\,m ahead of the robot, at the same place for every attempt, and the scene is reset before each one.
The robot finds the apple with the segmenter and localizes by its own odometry, as on the real robot.
A session is \rlWarmup\ warm-up attempts and 45 more.
A coin assigns each of these our learner's correction, no correction or the correction with exploration, with probability 1/3 each; for the estimators it assigns their correction or none, with probability 1/2 each.
Each learner runs three sessions.
\Cref{fig:rl}a plots the median end-effector error over the hold, pooled over the sessions in bins of five attempts; the control curve pools the attempts assigned no correction in the nine sessions of our learner and the estimators.
The sessions of \cref{fig:suppl_ablation} follow the same protocol.

\paragraph{Random placements (\cref{sec:ablation})}
Each placement puts the object on the floor 0.30 to 0.50\,m ahead of the pelvis and 0.02 to 0.18\,m to its left, turned by up to $30^\circ$ either way.
Placements are drawn from the session's seed, so every learner meets the same ones.
The grasp target and the robot's pose are read from the simulator's exact state, and the robot does not walk.
A session is \rlWarmup\ warm-up placements of one attempt each, then ten scored placements, and the learner carries over from one placement to the next.
A miss is retried in place, and the scene is restored to the placement after five misses in a row or when the object leaves the placement area.
A placement ends at the first lift or after ten attempts.
Each learner and the control run three sessions per object.
\Cref{sec:ablation} reports the shares of scored placements lifted at the first attempt and within ten attempts, with the three objects weighted equally.

\paragraph{Object height (\cref{fig:rl}c)}
The object stands on a static block whose top is the given height above the floor, 12\,cm square under the apple and 25\,cm square under the can; at 0\,cm the apple lies on the floor.
The floor pick takes the apple with the three-finger end effector at 0, 5, 10, 20, 30, 40, 50, 60, 70 and 80\,cm, and the standing pick takes a 330\,ml can with the jaw gripper at 20, 30, 40, 45, 50, 60, 70 and 80\,cm.
Each height is one session of 15 attempts with a fresh learner: the first \rlWarmup\ are the fast system's warm-up and the other ten are scored.
The robot finds the object with the head camera and the segmenter and localizes from the simulator's root pose; the fast system acts forward and sideways.
The harness adapts the height as in \cref{sec:edit}, with the knees also kept at their demonstrated heights.

%% file: arxiv/sec/suppl/protocol.tex
\section{Real-robot protocol and hardware}
\label{sec:suppl_protocol}

\paragraph{Robot and sensing}
The real-robot experiments (\cref{tab:real_ablation}) run on a \robotDoF-DoF Unitree G1 with a Dex3 three-finger end effector on the right wrist.
All sensing is onboard: the built-in Intel RealSense RGB-D head camera and the joint state, which gives $c_t$ by forward kinematics.

\paragraph{What runs where}
The robot's Jetson Orin NX runs the \controlRateHz\,Hz control loop, in which the whole-body motion tracker takes the harness's command plus the correction as its end-effector command, and beside it the actor, which computes the correction of \cref{eq:actor} at every control step, together with the end-effector driver and the retry.
A PC on the local network runs the VLM agent, the segmenter~\cite{sam3_2025}, which reads the head camera at a fraction of the control rate, and the learner's updates, which run in the background and hand the newest actor to the robot.

\paragraph{Attempts, episodes and sessions}
An attempt is one pass through the task's steps (\cref{sec:method}); it ends at the end state, at a fall, at a \timeoutMin-minute budget on any phase, or at a miss the robot detects itself (\cref{sec:retry}) and retries in place.
An episode is the attempts on one placement of the object, up to its end state, cut short by a fall, by an object never seen from the robot's stance or by the cap of ten attempts.
A session is one launch of the robot's program; each task step has its own learner, which starts from zero at the step's first entry and is kept across the session's episodes.

\paragraph{Learner and control}
Each learner and the control run sessions of their own, and a learner's \rlWarmup\ warm-up attempts are not scored.
An attempt succeeds when the robot stands up holding the object, by its own lift check, confirmed by the operator.

\paragraph{Placements (\cref{fig:rl}b)}
A placement runs from the operator placing the object to a lift; the eligible placements of \cref{fig:rl}b lie within the scored attempts of \cref{tab:real_ablation}, except any begun before them or cut off before their fifth try.

\paragraph{Labels and endpoints}
For the slow system's offline evaluation (\cref{tab:slow,tab:suppl_slow_indicator}), Claude Opus 5.5 labels every saved frame from the task's words alone: whether the object is in view and where, and for an attempt's frames whether the end effector holds it.
It never sees a phrase or a condition. Its labels of what the end effector holds are right on 92 of 105 simulated attempts, judged against the simulator's state.
A second labeler, \judgeModel, agrees with it on visibility on 582 of 607 frames ($\kappa=0.70$) and on 181 of the 186 held-out frames of \cref{tab:slow} ($\kappa=0.63$).
Every phrase comparison keeps its direction on the frames both labelers agree on.
Opus thus selects by the labels that score it, while Sonnet and Haiku select by their own.
The end-effector error, $\lVert c_t-n_t\rVert$ at contact, takes $c_t$ from the joint state and the target $n_t$, both in the attempt's frame, and needs no label.

\paragraph{What a person does}
No one teleoperates the robot.
Within an episode no one touches or resets it, and a person only sets the object back in view when it leaves the camera's field of view; after a success, a person places the object for the next episode.